%% file: main.tex
\documentclass{article}

\usepackage{aistats2019v2}

\usepackage[utf8]{inputenc} 
\usepackage[T1]{fontenc}    
\usepackage{hyperref}       
\usepackage{url}            
\usepackage{booktabs}       
\usepackage{amsfonts}       
\usepackage{nicefrac}       
\usepackage{microtype}      

\usepackage{blindtext}

\usepackage{amsthm}
\usepackage{amsfonts}
\usepackage{amssymb}
\usepackage{amsmath}
\usepackage{bbm}

\usepackage[ruled,vlined]{algorithm2e}

\usepackage{graphicx}
\usepackage{threeparttable}

\usepackage{caption}
\usepackage{subcaption}
\usepackage{tabu}
\DeclareMathOperator*{\argminA}{arg\,min}

\newcommand\smallO{
  \mathchoice
    {{\scriptstyle\mathcal{O}}}
    {{\scriptstyle\mathcal{O}}}
    {{\scriptscriptstyle\mathcal{O}}}
    {\scalebox{.7}{$\scriptscriptstyle\mathcal{O}$}}
  }

\DeclareMathOperator{\aalpha}{\frac{\alpha_{t}}{2(1-\beta_{1_{t}})}}

\newcommand*{\Resize}[2]{\resizebox{#1}{!}{$#2$}}%

\title{Double Adaptive Stochastic Gradient Optimization}

\author{
    \quad Kin Gutierrez\textsuperscript{1} \quad Jin Li\textsuperscript{1} \quad Cristian Challu\textsuperscript{2} \quad Artur Dubrawski\textsuperscript{1}\\
    \textsuperscript{1}Carnegie Mellon University \quad \textsuperscript{2}ITAM \\
    \texttt{\{kdgutier, jinl2, awd\}@cs.cmu.edu } \texttt{cristian.challu@itam.mx} \\
}

\begin{document}

\maketitle
\begin{abstract}
Adaptive moment methods have been remarkably successful in deep learning optimization, particularly in the presence of noisy and/or sparse gradients. 
We further the advantages of adaptive moment techniques by proposing a family of double adaptive stochastic gradient methods~\textsc{DASGrad}. 
They leverage the complementary ideas of the adaptive moment algorithms widely used by deep learning community, and recent advances in adaptive probabilistic algorithms.
We analyze the theoretical convergence improvements of our approach in a stochastic convex optimization setting, and provide empirical validation of our findings with convex and non convex objectives. 
We observe that the benefits of~\textsc{DASGrad} increase with the model complexity and variability of the gradients, and we explore the resulting utility in extensions of distribution-matching multitask learning. 
\end{abstract}

\section{Introduction and Motivation}
\input{sections/section1_introduction.tex}

\section{Adaptive Gradient Methods} \label{background}
\input{sections/section2_background.tex}

\newpage
\section{Convergence Analysis} \label{convergence}
\input{sections/section3_convergence.tex}

\section{Implementation of Double Adaptive Stochastic Gradient Descent} \label{implementation}
\input{sections/section4_implementation.tex}

\section{Algorithm Comparison} \label{comparison}
\input{sections/section5_comparison.tex}

\section{Discussion} \label{discussion}
\input{sections/section6_discussion.tex}

\section{Conclusion} \label{conclusion}
\input{sections/section7_conclusion.tex}

\newpage\phantom{new}
\newpage
\bibliographystyle{plain}
\bibliography{citations.bib}

\newpage
\section*{Appendix}
\input{sections/section_appendix.tex}

\end{document}

%% file: sections/section1_introduction.tex
Stochastic gradient descent (SGD) is a widely used optimization method, and currently through backpropagation this algorithm has propelled the success of many deep learning applications. Duchi et al.\ triggered the interest in adaptive algorithms, with their \textsc{ADAGrad} method they adjusted the classical SGD algorithm and improved its performance in settings with sparse features or noisy gradients in general \cite{Duchi:2011:ASM:1953048.2021068}. Following \textsc{ADAGrad} many variants were proposed to make the optimization algorithm better suited for high dimensional objectives by dealing with the problem of rapid learning rate decay with \textsc{ADADelta}, \textsc{RMSProp}, \textsc{ADAM} and most recently \textsc{AMSGrad} \cite{DBLP:journals/corr/abs-1212-5701, Tieleman2012, DBLP:journals/corr/KingmaB14, j.2018on}. All these adaptive moment methods relied on the efficient use of the information of the geometry of the problem to improve the rate of convergence. 

The access to large datasets has posed challenges to optimization methods. For the case of gradient descent algorithms, the calculation of the complete gradient has become computationally challenging, leading to the use of its stochastic versions. The most common stochastic alternative is uniform sampling, while other stochastic strategies include fixed sampling and gradient based sampling \cite{Zhu:2018arXiv180300841Z, pmlr-v48-gopal16,adaptive_variance_reducing_for_SGD, bottou2016optimization}. The core idea behind adaptive probabilities methods is to improve the efficiency on the use of the information of the gradients, by minimizing their variance to improve the rate of convergence. The adaptive probabilities approach has mainly focused on traditional convex objectives, not representative of all the complexity of the objective functions currently used, as shown by some of the most recent work, around variants of stochastic dual coordinate ascent \cite{dual_coordinate_ascent, safe_adaptive_importance}. 

Motivated by the information usage efficiency of adaptive probabilities and the benefits shown by adaptive moments methods against challenging objective functions, we introduce our novel alternative double adaptive stochastic gradient algorithm, built with the combination of the complementary ideas from the adaptive moments and adaptive probabilities methods. We will refer to our proposed algorithm as \textsc{DASGrad} for Double Adaptive Stochastic Gradient optimization.




It should be noted that a small number of methods already explore variations of double adaptation in deep learning literature, but its use is limited to the scope of their specific applications~\cite{prioritized_experience_replay}, and they are not supported by rigorous theoretical analysis. In this paper we seize the opportunity to analyze the double adaptive algorithms by showing theoretical improvement guarantees and validating these improvements empirically. 
We explore in detail the conditions that enhance~\textsc{DASGrad} convergence improvements and demonstrate~\textsc{DASGrad}'s generalization properties extending it to multi-task learning.


%% file: sections/section2_background.tex
\textbf{Notation.}
In order to facilitate the proofs and reading process we introduce some simplified notation that will be common to the analyzed algorithms. Let $a, b\in \mathbb{R}^d$ and $M \in S^{d}_{+}$, then the multiplication of vector $a$ by the inverse of $M$ will be $M^{-1}a = a/M$. Let $\sqrt{a}$ be the element-wise square root of vector a, $a^2$ the element-wise square, $a/b$ the element-wise division, and $\text{max}(a,b)$ the element-wise max of vector a and vector b. Finally for any natural $n$ the set $\{1,\dots, n\}$ is denoted as $[n]$.

Let $\mathcal{T} = \{(x_i,y_i)\}^{n}_{i=1}$ be a training set; let $f: \Theta \times X \times Y \to \mathbb{R}$ be a differentiable function that represents the empirical risk of an agent over $\mathcal{T}$ for the parameters $\theta \in \Theta$, with $\Theta \subseteq \mathbb{R}^{d}$ a convex feasible set of parameters; let $S^{d}_{+}$ the set of positive definite matrices in $\mathbb{R}^{d\times d}$, for a given matrix $M \in S^{d}_{+}$ and parameter $\theta'\in \Theta$; let $\Pi_{\Theta, M}$ be projection operator defined by $\Pi_{\Theta, M}(\theta') = \argminA_{\theta \in \Theta} ||M^{1/2}(\theta-\theta')||$, which can be seen as regularization for Machine Learning purposes.

For the iterative stochastic optimization algorithm $\mathcal{A}$, let $i_{t}$ be a sampled index $i$ at step $t$ drawn from the training set indices $[n]$, with $i_{t} \sim p_{t}$ and  $p_{t} \in \Delta^{n}_{+} = \{ p \in \mathbb{R}^n \,:\, p_{i} > 0 \quad \Sigma_{i}p_{i}=1 \} $. We denote the evaluated risk $f(\theta,x_{i},y_{i})=f_{i}(\theta)$, the complete gradient $\nabla f(\theta_{t}) = \frac{1}{n} \Sigma_{i_{t}} \nabla f_{i_{t}}(\theta_{t})$ and the stochastic gradient $\nabla f_{i_{t}}(\theta_{t})$, analogous a full descent direction $m_{t} = \frac{1}{n} \Sigma_{i_{t}} m_{i_{t}}$ and a stochastic descent direction $m_{i_{t}}$. \\

\textbf{Stochastic Optimization Framework.}
To analyze the convergence of the stochastic optimization algorithm $\mathcal{A}$ we use the convex optimization setting where we assume that the objective function is convex with bounded gradients, that is $||\nabla f_{i} (\theta) ||_{\infty} \leq G$ for all $i \in [n], \; \theta \in \Theta$, and finally the parameter space $\Theta$ has bounded diameter, that is $||\theta-\theta'||_{\infty} \leq D$ for all $\theta,\theta' \in \Theta$.

For our purposes, the algorithm $\mathcal{A}$ at time $t$ chooses a distribution over the training set $\hat{p}_{t} \in \Delta^{n}_{+}$, obtains a training example $i_{t} \sim \hat{p}_{t}$ and its importance weights $\hat{w}_{i_{t}} = (1/n)/p_{i_{t}}$, then updates its parameters $\theta_{t} \in \Theta$ using the available data at time $t$ and the importance weights $\hat{w}_{i_{t}}$ to unbias the direction of the gradients. After the update, the algorithm incurs in a loss from an unknown function $f(\theta_{t})$. To assess the performance of the algorithm after $T$ steps we use the expected regret, which measures the difference of the loss at time $t$ and the loss for optimal fixed parameter, along the possible trajectories induced by the chosen probabilities.

\begin{center}{ $R(\mathcal{A}) = \sum^{T}_{t=1} \mathbb{E}_{n} \left[\; f_{i}(\theta_{t}) - \text{min}_{\theta }\mathbb{E}_{n}[f_{i}(\theta)] \;\right] $ }\end{center}


\newpage
The goal is to design an algorithm $\mathcal{A}$ that has sub linear expected regret $R(\mathcal{A})_{T} = \smallO (T)$, which in turn implies that the algorithm will converge on average to the optimal parameter.

\begin{algorithm}[h]
\begin{footnotesize}
\caption{{\bf General Stochastic Gradient Method} \label{Algorithm1}}
\KwIn{$\theta_{1} \in \Theta$, step size $\{\alpha_{t}>0\}^{T}_{t=1}$, functions $\{\phi_{t}, \psi_{t}\}^{T}_{t=1}$} 
\For{$t=1$ \KwTo $T$}{
    Choose $\hat{p}_{t} \in \Delta^{n}_{+}$, and sample $i_{t} \sim \hat{p}_{t}$ \\
    Calculate $g_{i_{t}} = \nabla f_{i_{t}}(\theta_{t})$  and $\hat{w}_{i_{t}} = (1/n)/\hat{p}_{i_{t}}$ \\
    $m_{i_{t}} = \phi_{t}\left(g_{i_{1}},\dots,g_{i_{t}}\right) $ and $\hat{V}_{i_{t}} = \psi_{t}\left(g_{i_{1}},\dots,g_{i_{t}}\right)$ \\
    $\hat{\theta}_{t+1} = \theta_{t} - \alpha_{t} \hat{w}_{i_{t}} \, m_{i_{t}}/\sqrt{\hat{V}_{i_{t}}}$ \\
    $\theta_{t+1} = \Pi_{\Theta, \sqrt{\hat{V}_{i_{t}}}}(\hat{\theta}_{t+1})$
    }
\end{footnotesize}
\end{algorithm}

\textbf{General Stochastic Gradient Method.}

Algorithm \ref{Algorithm1} constitutes a general family of line search methods. This algorithm comprehends the classical stochastic gradient descent, adaptive methods family, and Newton methods \cite{Duchi:2011:ASM:1953048.2021068} \cite{NoceWrig06}, as we can obtain first and second order stochastic line search methods, varying the averaging functions of the past gradients with $\phi_{t}: \Theta^{t} \to \mathbb{R}^{d}$, and approximating the Hessian with the functions $\psi_{t}: \Theta^{t} \to S^{d}_{+}$.

\textbf{Adaptive Probabilities Methods.}
\input{sections/section2_1_adaptive_probabilities.tex}

\newpage
\textbf{Adaptive Moments Methods.}
\input{sections/section2_2_adaptive_moments.tex}

\textbf{Double Adaptive Methods.}
\input{sections/section2_3_double_adaptive.tex}

%% file: sections/section2_1_adaptive_probabilities.tex
The stochastic gradient descent algorithm is recovered with the following step size, sampling probabilities and functions: 
\begin{equation}
  \tag{\textsc{SGD}}
  \begin{split}
  \alpha_t = \alpha/\sqrt{t} \quad p_{i_{t}} = 1/n \; \text{ for all } t \in [T], i\in [n] \\
  \phi_t(g_{i_{1}},\dots,g_{i_{t}}) = g_{i_{t}} \; \psi_{t}(g_{i_{1}},\dots,g_{i_{t}}) = \mathbb{I} 
  \end{split}
  \label{eqn:SGD}
\end{equation}

Adaptive probabilities methods can be obtained simply by allowing the algorithm to choose a different probability $\hat{p}_{t}$ at any time $t$:
\begin{equation}
  \tag{\textsc{ap-SGD}}
  \begin{split}
  \alpha_t = \alpha/\sqrt{t} \quad \hat{p}_{t} \in \Delta^{n}_{+} \; \text{ for all } t \in [T] \\
  \phi_t(g_{i_{1}},\dots,g_{i_{t}}) = g_{i_{t}} \; \psi_{t}(g_{i_{1}},\dots,g_{i_{t}}) = \mathbb{I} 
  \end{split}
  \label{eqn:as-SGD}
\end{equation}

Significant improvements in the convergence rate of the algorithm can be obtained by cleverly choosing and computing such probabilities that in turn enables the algorithm to use data in a more efficient manner \cite{safe_adaptive_importance}. Fixed importance sampling is the special case when $\hat{p}_{t} = p$ for all $t \in [T]$.

%% file: sections/section2_2_adaptive_moments.tex
Duchi et al. triggered interest and research on adaptive algorithms. In their work they noticed that SGD lacked good convergence behavior in sparse settings, and proposed a family of algorithms that allowed the methods to dynamically incorporate information about the geometry of the data \cite{Duchi:2011:ASM:1953048.2021068}. Following huge gains obtained with \textsc{ADAGrad}, the deep learning community proposed variants based on exponential moving average functions for $\psi_{t}$ like \textsc{ADADelta}, \textsc{RMSProp}, \textsc{ADAM} and most recently \textsc{AMSGrad} \cite{DBLP:journals/corr/abs-1212-5701, Tieleman2012, DBLP:journals/corr/KingmaB14, j.2018on}.

The first algorithm \textsc{ADAGrad} is obtained by the following proximal functions:

\begin{equation}
  \tag{\textsc{ADAGrad}}
  \begin{split}
  \alpha_t = 1/\sqrt{t} \quad p_{i_{t}} = 1/n \; \text{ for all } t \in [T], i\in [n] \\
  \phi_t(g_{i_{1}},\dots,g_{i_{t}}) = g_{i_{t}} \\ \psi_{t}(g_{i_{1}},\dots,g_{i_{t}}) = \frac{1}{t} \text{diag}(\Sigma^{t}_{\tau=1} g^{2}_{i_{\tau}})
  \end{split}
  \label{eqn:ADAGRAD}
\end{equation}

The algorithm \textsc{AMSGrad} is obtained by setting:

\begin{equation}
  \tag{\textsc{AMSGrad/Adam}}
  \begin{split}
  \alpha_t = 1/\sqrt{t} \quad p_{i_{t}} = 1/n \; \forall t \in [T], i\in [n] \\
  \phi_t(g_{i_{1}},\dots,g_{i_{t}}) = \Sigma^{t}_{\tau=1} \text{\footnotesize $\beta_{1}(t)_{\tau}$} g_{i_{\tau}} \\
  v_{i_{t}} = (1-\text{\footnotesize $\beta_{2}$})\Sigma^{t}_{\tau=1} \text{\footnotesize $\beta^{t-\tau}_{2}$} g^{2}_{i_{\tau}} \quad \hat{v}_{i_{t}} = \text{max}(\hat{v}_{i_{t-1}}, v_{i_{t}}) \\
  \psi_{t}(g_{i_{1}},\dots,g_{i_{t}}) = \text{diag}\left(\hat{v}_{t}\right)
  \end{split}
  \label{eqn:ADAM}
\end{equation}

Fortunately a very simple and computationally efficient way to implement \textsc{AMSGrad} is given by a recursion. \textsc{RMSProp} is the particular case of \textsc{AMSGrad} when $\beta_1 = 0$ and without maximum operator for the second moments vector, while \textsc{ADAM} is recovered without the maximum operator. As was shown by Duchi et al. the expected regret can achieve an upper bound much better than $\mathcal{O}(\sqrt{dT})$ when in the sparse setting.


%% file: sections/section2_3_double_adaptive.tex
The key idea behind both the adaptive probabilities methods and adaptive moment methods is the efficient use of the information available in the training data to improve the performance of the algorithms. In the case of adaptive sampling methods, the probabilities $\hat{p}_{i_{t}}$, updated dynamically, use the information of the gradients to improve the convergence rate, while adaptive moment methods use information about the geometry of the problem.

For the analysis we will refer to double adaptive stochastic gradient algorithms from the general framework provided by  Algorithm \ref{Algorithm1}, built with the complementary ideas from the adaptive moments and adaptive probabilities methods as \textsc{DASGrad}.


%% file: sections/section3_convergence.tex
\newtheorem{theorem}{Theorem}
\newtheorem{corollary}{Corollary}[theorem]
\newtheorem{lemma}{Lemma}

In this section we provide the expected regret guarantees for common versions of Algorithm \ref{Algorithm1}. All the proofs of the theorems and corollaries are included in the Appendix.

\subsection{Convergence of Stochastic Gradient Descent}

Under  expected regret is the obtained from the greedy projection, adapted to the stochastic case with infinity norm bounds \cite{Zinkevich:2003:OCP:3041838.3041955}.

\begin{theorem}
\label{sgd_convergence}
Let $\{\theta_{t}\}^{T}_{t=1}$ be the sequence obtained with \textsc{SGD} then the expected regret bound is:
\[\begin{split}
R(\Resize{0.6cm}{\textsc{SGD}}) \leq 
\sum^{T}_{t=1}\frac{\alpha_{t}}{2}
\mathbb{E}_{n} \left[\, ||g_{i_{t}}||^{2}_{2}\,\Big|\,\theta_{t}\,\right] + \quad \quad \quad \quad \quad \quad \quad \\
\sum^{T}_{t=1} \frac{1}{2\alpha_{t}} \mathbb{E}_{n} \left[\,||\theta_{t}-\theta^{*}||^{2}_{2} - \mathbb{E}_{n}[\,||\theta_{t+1}-\theta^{*}||^{2}_{2}\,|\,\theta_{t}]\,\right] 
\end{split}
\]
\end{theorem}

\begin{corollary}
\label{sgd_corollary}
Following the sequence $\{\theta_{t}\}^{T}_{t=1}$ of \textsc{SGD} with step size $\alpha_{t} = 1 /\sqrt{t}$ and uniform probabilities $p_{i_{t}} = 1/n$, if we assume that $\Theta$ has bounded diameter $D$ and $||\nabla f_{i_{t}} (\theta)||_{\infty} \leq G$ for all $t \in [T]$ and $\theta \in \Theta$, then the expected regret bound is:
\[
R( \Resize{0.6cm}{\textsc{SGD}}) \leq d G^{2} (\sqrt{T}-1/2) + \frac{d D^{2}}{2} \sqrt{T} 
\]
\end{corollary}

\subsection{Convergence of Adaptive Probabilities Stochastic Gradient Descent}

We prove that in the case of adaptive probabilities we get the following improved bounds.

\begin{theorem}
\label{apsgd_convergence}
Let $\{\theta_{t}\}^{T}_{t=1}$ be the sequence obtained with \textsc{ap-SGD}, then the expected regret bound for any trajectory of probabilities $p_{t} \in \Delta^{n}_{+}$ is:
\[\begin{split}
R( \Resize{1cm}{\textsc{ap-SGD}}) \leq  
\sum^{T}_{t=1}\frac{\alpha_{t}}{2} 
\mathbb{E}_{p_{t}} \left[\,w^{2}_{i_{t}} ||g_{i_{t}}||^{2}_{2}\,\Big|\,\theta_{t}\,\right] + \quad \quad \quad \quad \quad \quad \\
\quad \sum^{T}_{t=1} \frac{1}{2\alpha_{t}} \mathbb{E}_{p_{1:t-1}} \left[\,||\theta_{t}-\theta^{*}||^{2}_{2} - \mathbb{E}_{p_{t}}[\;||\theta_{t+1}-\theta^{*}||^{2}_{2}\,|\,\theta_{t}]\;\right]
\end{split}
\]
\end{theorem}

\begin{corollary}
\label{apsgd_corollary}
Following the sequence $\{\theta_{t}\}^{T}_{t=1}$ of \textsc{ap-SGD}, with step size $\alpha_{t} = \alpha/\sqrt{t}$ and optimal adaptive probabilities $\hat{p}_{i_{t}} \propto ||\nabla f_{i_{t}}(\theta_{t})||_{2}$, if we assume that $\Theta$ has bounded diameter $D$ and $||\nabla f_{i_{t}} (\theta)||_{\infty} \leq G$ for all $t \in [T]$ and $\theta \in \Theta$, then the expected regret bound is:
\[\begin{split}
R( \Resize{1cm}{\textsc{ap-SGD}}) \leq 
d G^{2} (\sqrt{T}-1/2) 
\quad \quad \quad \quad \quad \quad \quad \\
- \sum^{T}_{t=1} \text{Var}_{n}\left(||\nabla f_{i_{t}}(\theta_{t})||_{2}\right)
+ \frac{d D^{2}}{2} \sqrt{T} 
\end{split}
\]
\end{corollary}


\subsection{Convergence of Adaptive Moments Methods}

We provide a bound for the expected regret of \textsc{AMSGrad} adapted to the stochastic case following the arguments in Reddi et al, Kingma $\&$ Ba \cite{j.2018on} \cite{DBLP:journals/corr/KingmaB14}. \\
\begin{theorem}
\label{amsgrad_convergence}
Let $\{\theta_{t}\}^{T}_{t=1}$ be the sequence obtained with \textsc{AMSGrad}, then the regret bound is:
\[\begin{split}
R( \Resize{1.4cm}{\textsc{AMSGrad}}) \leq  
\sum^{T}_{t=1}\frac{\alpha_{t}}{2(1-\beta_{1_{t}})} \mathbb{E}_{n} \left[ ||\hat{V}^{-1/4}_{i_{t}}m_{i_{t}}||^{2}_{2}\right] 
\quad \quad \quad \quad \quad \quad \quad \\
\, +\sum^{T}_{t=1} \frac{1}{2\alpha_{t}(1-\beta_{1_{t}})} \mathbb{E}_{n} \left[||\hat{V}^{1/4}_{i_{t}}(\theta_{t}-\theta^{*})||^{2}_{2}\right] \quad \quad \quad \quad \quad \quad \quad\\ -\frac{1}{2\alpha_{t}(1-\beta_{1_{t}})} \mathbb{E}_{n} \left[ ||\hat{V}^{1/4}_{i_{t}}(\theta_{t+1}-\theta^{*})||^{2}_{2}\right] \quad \quad \quad \quad  \quad \quad \; \\ 
+ \sum^{T}_{t=1} \frac{\alpha_{t} \beta_{1_{t}}}{2(1-\beta_{1_{t}})} ||\hat{V}^{-1/4}_{t}m_{t-1}||^{2}_{2} 
\quad \quad \quad \quad \quad  \quad \quad \\
+ \sum^{T}_{t=1} \frac{\beta_{1_{t}}}{2\alpha_{t}(1-\beta_{1_{t}})}
||\hat{V}^{1/4}_{t}(\theta_{t}-\theta^{*})||^{2}_{2} 
\quad \quad \quad \quad \quad  \quad \quad 
\end{split}
\]
\end{theorem}

\begin{corollary}
\label{amsgrad_corollary}
Following the sequence $\{\theta_{t}\}^{T}_{t=1}$ of \textsc{AMSGrad} with step size $\alpha_{t} = \alpha/\sqrt{t}$, averaging parameters $\beta_{1} = \beta_{1_{1}}$, $\beta_{1_{t}} \leq \beta_{1}$ for all $t\in[T]$, $\gamma = \beta_{1}/\sqrt{\beta_2} < 1$ and uniform probabilities $p_{i_{t}} = 1/n$. If we assume that $\Theta$ has bounded diameter $D$, $||\nabla f_{i_{t}} (\theta)||_{\infty} \leq G$ for all $t \in [T]$ and $\theta \in \Theta$, then the expected regret bound is:
\[\begin{split}
R( \Resize{1.4cm}{\textsc{AMSGrad}}) \leq  
\quad \quad \quad \quad \quad \quad \quad 
\quad \quad \quad \quad \quad \quad \quad 
\quad \quad \quad \quad \quad \quad \quad 
\\
\; \frac{D^{2}\sqrt{T}}{2\alpha(1-\beta_{1})}
\frac{\alpha \sqrt{1+\log(T)}}{2(1-\beta_{1})^2\sqrt{(1-\beta_2)}(1-\gamma)} \sum^{d}_{h=1} ||\;\bar{|\,g\,|}_{1:T,h}\;||_{2}
\quad \quad \quad \\
+\mathbb{E}_{n}\left[||\hat{v}^{1/4}_{i_{T}}||^{2}_{2}\right] 
+ \frac{\alpha G d}{2\alpha(1-\beta_{1})^{3} \sqrt{(1-\beta_{2})} (1-\gamma) } 
\quad \quad \quad \quad \\
+ \frac{D^{2}}{2\alpha(1-\beta_{1})} \sum^{T}_{t=1} \sqrt{t} \beta^{T-t}_{1} ||\hat{v}^{1/4}_{t}||^{2}_{2}
\quad \quad \quad \quad \\
\end{split}
\]
\end{corollary}

\subsection{Convergence of Double Adaptive Methods}

\begin{theorem}
\label{dasgrad_convergence}
Let $\{\theta_{t}\}^{T}_{t=1}$ be the sequence obtained with \textsc{DASGrad}, then the regret bound for any trajectory of probabilities $p_{t} \in \Delta^{n}_{+}$ is:
\[\begin{split}
R( \Resize{1.4cm}{\textsc{DASGrad}}) \leq
\sum^{T}_{t=1}\frac{\alpha_{t}}{2(1-\beta_{1_{t}})} \mathbb{E}_{p_{1:t}} \left[\,w^{2}_{i_{t}} ||\hat{V}^{-1/4}_{i_{t}}m_{i_{t}}||^{2}_{2}\,\right] + 
\quad \quad \quad \quad \quad \quad \\
\sum^{T}_{t=1} \frac{1}{2\alpha_{t}(1-\beta_{1_{t}})} \mathbb{E}_{p_{1:t}} \left[||\hat{V}^{1/4}_{i_{t}}(\theta_{t}-\theta^{*})||^{2}_{2} - ||\hat{V}^{1/4}_{i_{t}}(\theta_{t+1}-\theta^{*})||^{2}_{2}\right] 
\quad \quad \quad \\ 
+ \sum^{T}_{t=1} \frac{\alpha_{t} \beta_{1_{t}}}{2(1-\beta_{1_{t}})} ||\hat{V}^{-1/4}_{t}m_{i_{t-1}}||^{2}_{2}
\quad \quad \quad \quad \quad \\
+ \sum^{T}_{t=1} \frac{\beta_{1_{t}}}{2\alpha_{t}(1-\beta_{1_{t}})}
||\hat{V}^{1/4}_{t}(\theta_{t}-\theta^{*})||^{2}_{2}
\quad \quad \quad \quad \quad  \\
\end{split}
\]
\end{theorem}

\begin{corollary}
\label{dasgrad_corollary}
Following the sequence $\{\theta_{t}\}^{T}_{t=1}$ of \textsc{DASGrad}, step size $\alpha_{t} = \alpha/\sqrt{t}$, averaging parameters $\beta_{1} = \beta_{1_{1}}$, $\beta_{1_{t}} \leq \beta_{1}$ for all $t\in[T]$, $\gamma = \beta_{1}/\sqrt{\beta_2} < 1$ and the optimal adaptive probabilities $\hat{p}_{i_{t}} \propto ||\hat{V}^{-1/4}_{i_{t}}m_{i_{t}}||_{2}$. If we assume that $\Theta$ has bounded diameter $D$ and $||\nabla f_{i_{t}} (\theta)||_{\infty} \leq G$ for all $t \in [T]$ and $\theta \in \Theta$, then the expected regret bound is:
\[\begin{split}
R(\Resize{1.4cm}{\textsc{DASGrad}}) \leq  
\quad \quad \quad \quad \quad \quad \quad 
\quad \quad \quad \quad \quad \quad \quad 
\quad \quad \quad \quad \quad \quad \quad 
\\
\frac{\alpha \sqrt{1+\log(T)}}{2(1-\beta_{1})^2\sqrt{(1-\beta_2)}(1-\gamma)} \sum^{d}_{h=1} ||\;\bar{|\,g\,|}_{1:T,h}\;||_{2} 
\quad \quad \quad \quad \quad \\
-\sum^{T}_{t=1} \text{Var}_{n}\left(||\hat{V}^{1/4}_{i_{t}}m_{i_{t}}||_{2}\right)+\frac{D^{2}\sqrt{T}}{2\alpha(1-\beta_{1})} \mathbb{E}_{\hat{p}_{1:T}}\left[||\hat{v}^{1/4}_{i_{T}}||^{2}_{2}\right] 
\quad \quad \quad \quad \quad \\
+ \frac{\alpha G d}{2\alpha(1-\beta_{1})^{3} \sqrt{(1-\beta_{2})} (1-\gamma) } 
\quad \quad \quad \quad \quad  \\
+ \frac{D^{2}}{2\alpha(1-\beta_{1})} \sum^{T}_{t=1} \sqrt{t} \beta^{T-t}_{1} ||\hat{v}^{1/4}_{t}||^{2}_{2}
\quad \quad \quad \quad \quad  \\
\end{split}
\]
\end{corollary} 
With Corollaries \ref{apsgd_corollary} and \ref{dasgrad_corollary} we observe that \textsc{DASGrad} indeed improves the convergence rate over adaptive moment methods with uniform sampling, and these improvements increase with the variance of the gradients, so therefore problems involving more complex data and models will benefit more.


%% file: sections/section4_implementation.tex
As proven in Corollary 4.1, using optimal adaptive probabilities for sampling can in theory improve the convergence of the adaptive moments family. 

To update and sample we follow the common practice to address large multinomial distributions by using a segment tree data structure. It allows updating the distribution at $\mathcal{O}(n\log{}n)$ and sample from it at $\mathcal{O}(\log{}n)$. This tree data structure stores the adaptive probabilities of each training sample in its leafs and stores in each node the sum of the probabilities of the children. For an adaptive sampling algorithm to be practical, we must rely on approximations of the gradients. Due to that, we compute and update the optimal probabilities every $J$ iterations.\footnote{To enhance numerical stability, we add a small constant $\epsilon$ to each probability.} 

Based on the above ideas we propose to implement the Double Adaptive Stochastic Gradient method \textsc{DASGrad} using the pseudo-code in Algorithm \ref{Algorithm2}.

\begin{algorithm}[ht]
\begin{footnotesize}
\caption{{\bf \textsc{DASGrad}} \label{Algorithm2}}
\KwIn{$\theta_{1} \in \Theta$, functions $\{\phi_{t}, \psi_{t}\}^{T}_{t=1}$, frequency $J$} 
\For{$t=1$ \KwTo $T$}{
    \uIf{$t \; \text{mod} \; J = 0$}{
    Compute $\hat{p}_{t} \in \Delta^{n}_{+}$ setting 
    $\hat{p}_{i_{t}} \propto ||\hat{V}^{-1/4}_{i_{t}}m_{i_{t}}||_{2} +\epsilon$
    }
    Sample $i_{t} \sim \hat{p}_{t}$ using the segment tree \\
    Calculate $g_{i_{t}} = \nabla f_{i_{t}}(\theta_{t})$  and $\hat{w}_{i_{t}} = (1/n)/\hat{p}_{i_{t}}$ \\
    $m_{t} = \beta_{1t} m_{t-1} + (1 - \beta_{1t}) g_{t}$ and $v_{t} = \beta_{2}v_{t-1} + (1 - \beta_{2}) g^{2}_{t}$ \\
    $\hat{v}_{t} = max(\hat{v}_{t-1},v_{t})$ and $\hat{V}_{t} = diag(\hat{v}_{t})$ \\
    $\hat{\theta}_{t+1} = \theta_{t} - \alpha_{t} \hat{w}_{i_{t}} \, m_{i_{t}}/\sqrt{\hat{V}_{i_{t}}}$ \\
    $\theta_{t+1} = \Pi_{\Theta, \sqrt{\hat{V}_{i_{t}}}}(\hat{\theta}_{t+1})$
    }
\end{footnotesize}
\end{algorithm}

%% file: sections/section5_comparison.tex

Adaptive moment methods can outperform classical gradient descent methods by integrating the geometry of the problem with a diagonal approximation of the Hessian. It was shown by Duchi et al.\ that the adaptive moment methods can achieve an exponentially smaller bound for the expected regret with respect to the dimensionality of data $d$, when dealing with sparse features or small gradients in general \cite{Duchi:2011:ASM:1953048.2021068}. Based on the results from Theorem \ref{sgd_convergence} and Theorem \ref{amsgrad_convergence}, the expected regret bound of \textsc{SGD} is $\mathcal{O}(\sqrt{d T})$, while for the adaptive moment methods in the sparse setting, the potential and error component of the expected regret each will satisfy: 
\[\begin{split}
\mathbb{E}_{p_{1:T}}\left[||\hat{v}^{1/4}_{i_{T}}||^{2}_{2}\right] = \mathbb{E}_{p_{1:T}}\left[\sum^{d}_{h=1} \hat{v}^{1/2}_{i_{T},h}\right] \ll \sqrt{d} \\
\sum^{d}_{h=1} ||\;\bar{|\,g\,|}_{1:T,h}\;||_{2}
\ll \sqrt{d T}
\end{split}
\]
which in turn translates to a much better expected regret bound than $\mathcal{O}(\sqrt{d T})$.

Complementary to that, the adaptive probabilities methods can outperform SGD methods, because they allow the algorithm to re-evaluate the relative importance of each data point to maximize the expected learning progress, and minimize the variance of the stochastic gradient at each step.


To support the theoretical results, we provide empirical evidence that exhibits that with increased variance in the data, we have increased benefits of the double adaptive methods when compared to the state-of-the-art convergence rates. We demonstrate such relationship on classification problems using logistic regression and deep neural networks, comparing \textsc{Adam}, \textsc{AMSGrad}, and our \textsc{DASGrad}.

\textbf{Logistic Regression}: For the convex setting we solve two classification problems with L2 regularization. For the non sparse feature experiment we use the MNIST digit dataset, which is composed of $60,000$ images of $28\times 28$ hand written digits. For the sparse feature experiment we use the IMDB movie rating dataset which is composed of $25,000$ highly polar movie reviews and the sentiment label for the review \cite{Maas:2011:LWV:2002472.2002491}.\footnote{For both experiments, we use a batch of size 32, with a probability update every 10 steps, and the step size $\alpha_{t} = \alpha / \sqrt{t}$. We set $\beta_{1}=0.9$, $\beta_{2}=0.99$, and choose $\alpha$ through a grid search. For the MNIST dataset, for all three optimizers, the optimal learning rates are $\alpha=0.01$. For the IMDB dataset, we find the optimal learning rates to be $\alpha=0.005$ for \textsc{Adam}, $\alpha=0.006$ for \textsc{AMSGrad}, and $\alpha=0.02$ for \textsc{DASGrad}.}

\textbf{Neural Networks}: For the non convex setting we perform one experiment, we use the CIFAR10 dataset, which is composed of $60,000$ colour images of $32\times32$ pixels labeled in 10 classes. For this multiclass classification problem we use a convolutional neural network following the \textsc{small-CIFARNET} architecture, consisting of two convolution filters combined with max pooling and local response normalization, followed by two fully connected layers of rectified linear units \cite{Krizhevsky:2012:ICD:2999134.2999257}. \footnote{For the experiment we use a batch size of 32, with a probability update every 300 steps, and step size of $\alpha_{t} = \alpha / \sqrt{t}$. We set $\beta_{1}=0.9$, $\beta_{2}=0.99$, and choose $\alpha$ through a grid search, for which the optimal learning rate for all optimizers is $\alpha=0.001$.} 
\begin{figure}[ht]
\centering
\includegraphics[width=1\linewidth]{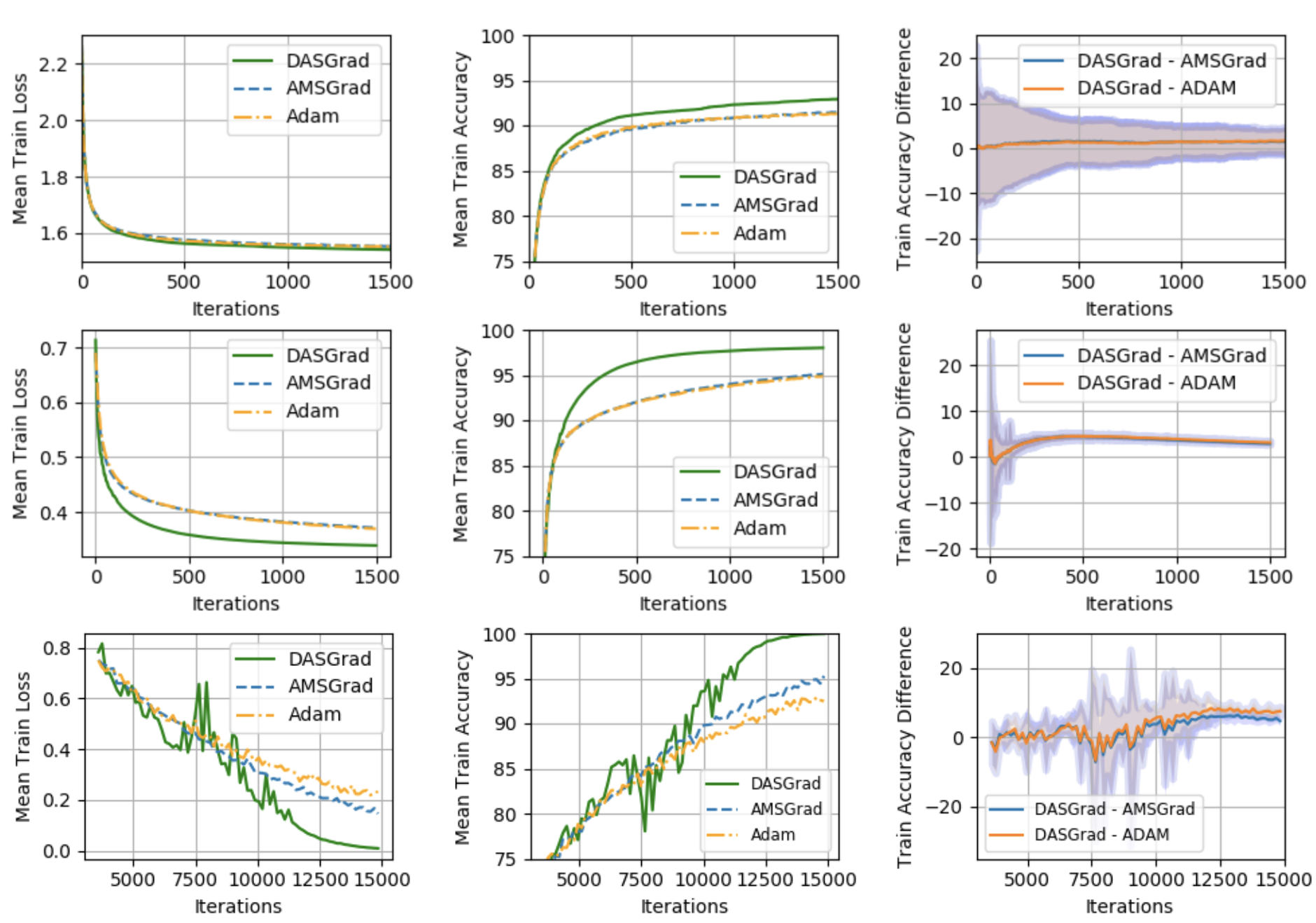}
\caption{Trajectories in convex and deep learning settings. First row 100 logistic regressions on MNIST, second row 100 logistic regressions on IMDB, third row 10 convolutional neural networks on CIFAR10. We show the mean over the trajectories for training loss (left), training accuracy (center), and the accuracy improvement of \textsc{DASGrad} with respect to \textsc{AMSGrad} and \textsc{Adam} with 95\% confidence intervals (right).} \label{fig:imdb_mnist_comparison}
\end{figure}

From the comparison in Figure \ref{fig:imdb_mnist_comparison}, we observe that in all cases the \textsc{DASGrad} optimization algorithm outperforms its adaptive moment counterparts represented by \textsc{Adam} and \textsc{AMSGrad}, as expected. The improvement is more significant for the IMDB dataset than it is for the MNIST dataset. From Figure \ref{fig:imdb_mnist_comparison} we can see that \textsc{DASGrad} continues to outperform \textsc{Adam} and \textsc{AMSGrad} in the deep learning setting. These results reinforce the previous statement that the benefits from DASGrad increase with the complexity of the data and the models. 

%% file: sections/section6_discussion.tex
\textbf{Improvements and variance.} To further explore the relationship between variance and the improvements to the convergence rate of the \textsc{DASGrad} algorithm, we implemented an online centroid learning experiment. Because of the linear relationship between the features and the gradients, we are able to explicitly control their variance. For this experiment, the empirical risk and gradients will be given by $R_n(\theta)=\frac{1}{2n}\sum_{i=1}^n||\theta-x_i||_2^2$ and $\nabla f(\theta, x_i)=\theta-x_i $.

\begin{figure}[ht]
\centering
        \includegraphics[width=0.75\linewidth]{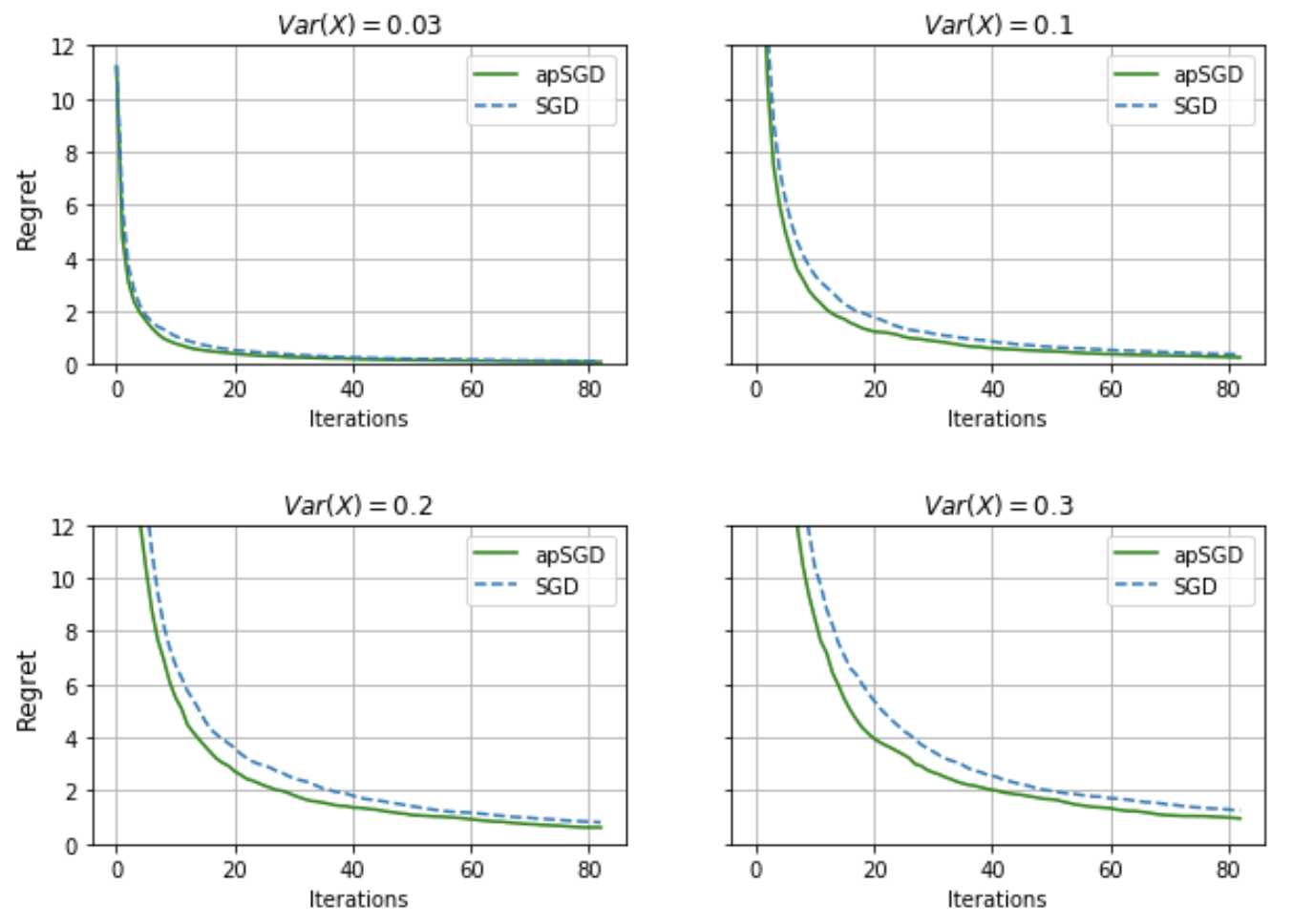}
        \caption{Trajectories of 100 random seeds, for the online centroid learning problem with different variance for the features. Enhanced improvements of adaptive methods with higher variance of the gradients.} \label{fig:mnist_variance}
\end{figure}

As we can see from Figure \ref{fig:mnist_variance} the greater the variance of the gradients, the greater the benefit that one can obtain from an adaptive probabilities method such as \textsc{DASGrad}, since those probabilities will prioritize the data points with the most learning potential.

\begin{figure}[ht]
\centering
        \includegraphics[width=01\linewidth]{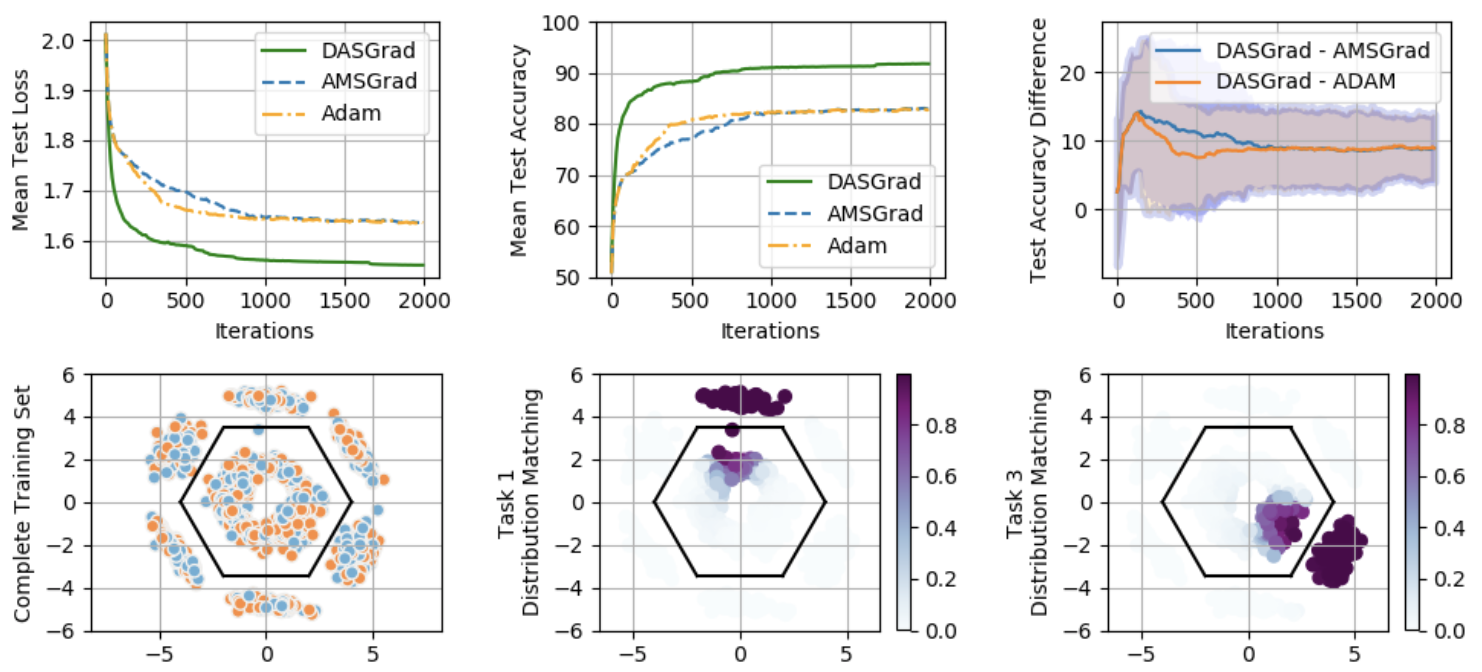} 
        \caption{Top row: trajectories of 20 random seeds for 2,000 iterations in convex optimization settings. Multiclass logistic regression on unbalanced MNIST dataset. We show the mean over 20 trajectories of training loss (left), training accuracy (center), and the improvement in accuracy of DASGrad with respect to \textsc{AMSGrad} and \textsc{Adam} with a 95\% confidence interval (right). Bottom row: example of multi-task learning for distribution matching through importance weights.} \label{fig:iw_shift}
\end{figure}

\textbf{Multitask Learning and Distribution Matching.} When the training $\mathcal{T}$ and test $\mathcal{T}'$ set do not share the same distribution, we may face a sample selection bias. Our \textsc{DASGrad} algorithm is compatible with the cost re-weighting correction technique \cite{Elkan:2001:FCL:1642194.1642224} as we can set the importance weights $w_{t}$ for any trajectory of distributions $p_{t}$, to unbias the gradients for the test distribution instead of the training.
\[
\begin{split}
R( \Resize{1.3cm}{\textsc{DASGrad}})_{\mathcal{T}'} 
= \sum^{T}_{t=1} \mathbb{E}_{p_{\mathcal{T}'}}\left[ f_{i}(\theta_{t})-\text{min}_{\theta }\mathbb{E}_{p_{\mathcal{T}'}}[f_{i}(\theta)]\right] \\
= \sum^{T}_{t=1} \mathbb{E}_{p_{1:t}}\left[w_{i_{t}}f_{i}(\theta_{t}) - \text{min}_{\theta }\mathbb{E}_{p_{\mathcal{T}'}}[f_{i}(\theta)]\right]
\end{split}
\]
To test the generalization properties of the \textsc{DASGrad} algorithm empirically, we unbalanced the MNIST training data set by reducing ninety percent the observations from the $1$ and $3$ digit. We set the importance weights to $w_{i_{t}}=(|L_{i}|/m)/p_{i_{t}}$, where $|L_{i}|$ is the count of the label $L$ associated with index $i$ in test over $m$, the number of test samples. As we see in Figure \ref{fig:iw_shift} using \textsc{DASGrad} with the correct importance weights has the desired generalization properties when facing a domain shift.

Another natural extension of \textsc{DASGrad} is towards multi-task learning. In a similar manner to transfer learning, we can use the importance weights to match the distributions for the estimator of the gradients to different tasks in the training set, similarly to \cite{Bickel:2009:DLU:1577069.1755858}. In Figure \ref{fig:iw_shift} we show an example of using importance weights for distribution matching.

\newpage
\textbf{Other Scenarios.} The applications of double adaptive stochastic gradient descend methods reach beyond supervised learning as shown by Schaul et al.~\cite{prioritized_experience_replay}. They improved performance of Deep Q-Network agents in reinforcement learning applications with the use of prioritized experience replay, using adaptive probabilities based on the temporal difference error.

In a more general sense, the double adaptive stochastic gradient algorithms satisfy the definition of meta-learning because we can use a learning subsystem to learn to spot the outliers on the gradient norms, and to help reduce the computational burden of computing the optimal probabilities. Such a model could also be pre-trained (as in 'Learning to Learn' approach \cite{DBLP:journals/corr/AndrychowiczDGH16}).

However, particularly when dealing with large datasets, the computational burden of calculating, updating and sampling from the optimal adaptive probabilities may counteract the attainable benefits, when compared to the uniform sampling. Therefore, adaptive probabilities methods require further exploration for efficient implementations, such as parallelizing the calculation of all gradients in the dataset.

%% file: sections/section7_conclusion.tex
Capability of learning from data efficiently is a prerequisite for practical success of complex learning models across various problem settings and application contexts. 
We have shown how to leverage the double adaptive stochastic gradient descent methodology
to enable efficient learning in a generalizable manner, while ensuring convergence improvement. 
We observed that our \textsc{DASGrad} algorithm outperforms currently prevalent variants of adaptive moment algorithms such as \textsc{Adam} and \textsc{AMSGrad} overall, in the context of the number of iterations required to achieve comparable performance, under the theoretical convergence guarantees in a stochastic convex optimization setting. With empirical validation in convex and non convex settings, we have shown that the advantages of the proposed algorithm become more prominent with the increasing complexity of data and models, and with more variance in the gradients. We have also broadened our results to demonstrate generalization properties of our approach and its extensions to multitask learning, as well as intuitive connections to other learning scenarios. 

%% file: sections/section_appendix.tex
\subsection*{A PROOF OF THEOREM \ref{apsgd_convergence}}
\input{sections/section_appendix_th2.tex}

\subsection*{B PROOF OF COROLLARY \ref{apsgd_corollary}}
\input{sections/section_appendix_c2_1.tex}

\subsection*{C PROOF OF THEOREM \ref{dasgrad_convergence}}
\input{sections/section_appendix_th4.tex}

\newpage
\subsection*{D PROOF OF COROLLARY \ref{dasgrad_corollary}}
\input{sections/section_appendix_c4_1.tex}


%% file: sections/section_appendix_th2.tex
The proof of Theorem \ref{apsgd_convergence} assumes a convex differentiable objective function $f$, bounded diameter for the parameters, and bounded norm of the gradients. And will be given for any trajectory of probabilities $p_{t}$, Corollary \ref{apsgd_corollary} addresses the optimal probabilities $\hat{p}_{t}$.

\begin{proof}
Since function $f$ is convex, for all $\theta$, the regret of period $t$ will be upper bounded by the product of the gradient $g_{t} = \nabla f (\theta_{t})$ and the difference $\theta_{t}$ and the fixed optimal $\theta^{*}$.
\[
f(\theta_{t}) - f(\theta^{*}) \leq \langle \,g_{t} ,\, \theta_{t}-\theta^{*} \rangle = \mathbb{E}_{n}\left[\langle \,g_{i_{t}} ,\, \theta_{t}-\theta^{*} \rangle\right]
\]

While using \textsc{ap-SGD} the parameter update will be given by the stochastically dependent on the observed training example $i_{t}$ and the current parameter $\theta_{t}$:
\[
\theta_{t+1} = \Pi_{\Theta, \mathbb{I}} \left(\theta_{t} - \alpha_{t} w_{i_{t}} g_{i_{t}}\right) 
= \argminA_{\theta \in \Theta} ||\theta_{t} - \alpha_{t} w_{i_{t}} g_{i_{t}}||_{2}
\]

Then to bound expected regret we use the fact that:
\[
\begin{split}
\hat{\theta}_{t+1}-\theta^{*} = (\theta_{t}-\theta^{*}) - \alpha_{t} w_{i_{t}} g_{i_{t}} 
\quad \quad \quad \quad \quad \quad \quad \quad \quad \\
||\hat{\theta}_{t+1}-\theta^{*}||^{2}_{2} = ||\theta_{t}-\theta^{*}||^{2}_{2} 
\quad \quad \quad \quad \quad \quad \quad \quad \quad \quad \quad \quad \\
- 2\alpha_{t} w_{i_{t}} \langle \,g_{i_{t}} ,\, \theta-\theta^{*} \rangle + \alpha^{2}_{t} w^{2}_{i_{t}} ||g_{i_{t}}||^{2}_{2}
\end{split}
\]

We identify the three components as the potential, the immediate cost and the error respectively. Taking the expectation at time $t$, and using the norm reduction property of the projections:
\[
\begin{split}
\mathbb{E}_{p_{t}}\left[\,||\theta_{t+1}-\theta^{*}||^{2}_{2} \,\Big|\, \theta_{t}\,\right] \leq 
||\theta_{t}-\theta^{*}||^{2}_{2} 
\quad \quad \quad \quad \quad \quad \quad \quad \quad \quad \; \\
- 2\alpha_{t}\mathbb{E}_{p_{t}}\left[w_{i_{t}} \langle \,g_{i_{t}} ,\, \theta_{t}-\theta^{*} \rangle \,\Big|\, \theta_{t}\,\right]
+ \alpha^{2}_{t} \mathbb{E}_{p_{t}} \left[\,w^{2}_{i_{t}} ||g_{i_{t}}||^{2}_{2}\,\Big|\,\theta_{t}\,\right] 
\quad \quad
\end{split}
\]

Since $w_{t}$ is the Radon-Nikodym derivative the interior product will be unbiased, then:
\[
\begin{split}
\mathbb{E}_{p_{t}}\left[\,||\theta_{t+1}-\theta^{*}||^{2}_{2} \,\Big|\, \theta_{t}\,\right] \leq 
||\theta_{t}-\theta^{*}||^{2}_{2} 
\quad \quad \quad \quad \quad \quad \quad \quad \quad \quad \; \\
- 2\alpha_{t} \mathbb{E}_{n}\left[\langle \,g_{i_{t}} ,\, \theta_{t}-\theta^{*} \rangle\right]
+ \alpha^{2}_{t} \, \mathbb{E}_{p_{t}} \left[\,w^{2}_{i_{t}} ||g_{i_{t}}||^{2}_{2}\,\Big|\,\theta_{t}\,\right] 
\quad \quad
\end{split}
\]

Rearranging the terms:
\[
\begin{split}
\langle \,g_{t} ,\, \theta_{t}-\theta^{*} \rangle \leq
\frac{\alpha_{t}}{2} \mathbb{E}_{p_{t}} \left[\,w^{2}_{i_{t}} ||g_{i_{t}}||^{2}_{2}\,\Big|\,\theta_{t}\,\right] 
\quad \quad \quad \quad \quad \quad \quad \\
+\frac{1}{2\alpha_{t}} \left(\,||\theta_{t}-\theta*||^{2}_{2} - \mathbb{E}_{p_{t}}\left[\,||\theta_{t+1}-\theta^{*}||^{2}_{2}\,\Big|\,\theta_{t}\,\right]\right)
\end{split}
\]

Finally summing until time $T$ and taking expectations of the complete trajectory of the algorithm:
\begin{equation}\begin{split}
\label{equation_apsgd_convergence}
R(\textsc{ap-SGD}) \leq  
\sum^{T}_{t=1}\frac{\alpha_{t}}{2} 
\mathbb{E}_{p_{t}} \left[\,w^{2}_{i_{t}} ||g_{i_{t}}||^{2}_{2}\,\Big|\,\theta_{t}\,\right] 
\quad \quad \quad \quad \quad \quad \quad \\
\sum^{T}_{t=1} \frac{1}{2\alpha_{t}} \mathbb{E}_{p_{1:t-1}} \left[\,||\theta_{t}-\theta^{*}||^{2}_{2} - \mathbb{E}_{p_{t}}[\,||\theta_{t+1}-\theta^{*}||^{2}_{2}\,|\,\theta_{t}]\,\right] 
\quad  \\ 
\end{split}
\end{equation} \\

\end{proof}

%% file: sections/section_appendix_c2_1.tex
\begin{proof}

Analogous to the proof of Theorem \ref{apsgd_convergence}, we demonstrate Lemma \ref{apsgd_convergence_lemma1} and Lemma \ref{apsgd_convergence_lemma3} following closely the convergence proof of the online greedy projection algorithm and adapt it to the stochastic case with infinity norm bounds. The proof of Corollary \ref{sgd_corollary} and \ref{apsgd_corollary} follows from the combination of the Lemmas.

Following the sequence $\{\theta_{t}\}^{T}_{t=1}$ of \textsc{ap-SGD} with step size $\alpha_{t} = \alpha/\sqrt{t}$, and bounded diameter $D$ for $\Theta$ and $||\nabla f_{i_{t}} (\theta)||_{\infty} \leq G$ for all $t \in [T]$ and $\theta \in \Theta$.

\begin{lemma}
\label{apsgd_convergence_lemma1}

From Equation \ref{equation_apsgd_convergence} the potential component will satisfy that:
\[
\begin{split}
\sum^{T}_{t=1} \frac{1}{2\alpha_{t}} \mathbb{E}_{p_{1:t-1}} \left[\,||\theta_{t}-\theta^{*}||^{2}_{2} - \mathbb{E}_{p_{t}}[\,||\theta_{t+1}-\theta^{*}||^{2}_{2}\,|\,\theta_{t}]\;\right] 
\quad \quad \\
\leq \frac{d D^{2}}{2} \sqrt{T} 
\quad \quad
\end{split}
\]

\textit{Proof}
\[
\begin{split}
\sum^{T}_{t=1} \frac{1}{2\alpha_{t}} \mathbb{E}_{p_{1:t-1}} \left[\,||\theta_{t}-\theta^{*}||^{2}_{2} - \mathbb{E}_{p_{t}}[\,||\theta_{t+1}-\theta^{*}||^{2}_{2}\,|\,\theta_{t}]\;\right]  
\quad \quad \quad \quad \quad \quad \quad \quad \quad \quad \quad \quad \quad \quad \quad \quad \quad \quad \quad \quad \quad \quad \quad \quad \quad \quad \quad \quad \quad  \quad\\
= \frac{1}{2\alpha_{1}} \mathbb{E}_{p_{1}}\left[||\theta_{1}-\theta^{*}||^{2}_{2}|\theta_{0}\right] -\frac{1}{2\alpha_{T}} \mathbb{E}_{p_{1:T}}\left[||\theta_{T+1}-\theta^{*}||^{2}_{2}|\theta_{T}\right]  
\quad \quad \quad \quad \quad \quad \quad \quad \quad \quad \quad \quad \quad \quad \quad \quad \quad \quad \quad \quad \quad \quad \quad \quad \quad \quad \quad \quad \; \\
+ \frac{1}{2}\sum^{T}_{t=2} \left(\frac{1}{\alpha_{t}}-\frac{1}{\alpha_{t-1}}\right)\mathbb{E}_{p_{1:t}}\left[\,||\theta_{t}-\theta^{*}||^{2}_{2}\,|\,\theta_{t-1}\right] 
\quad \quad \quad \quad \quad \quad \quad \quad \quad \quad \quad \quad \quad \quad \quad \quad \quad \quad \quad \quad \quad \quad \quad \quad \quad \quad \quad \quad \quad  \quad \\
\leq \frac{||D \odot \mathbbm{1}||^{2}_{2}}{2} \left(\frac{1}{\alpha_{1}} + \sum^{T}_{t=2} \left(\frac{1}{\alpha_{t}}-\frac{1}{\alpha_{t-1}}\right)\right) \quad \quad \quad \quad \quad \quad \quad \quad \quad \quad \quad \quad \quad \quad \quad \quad \quad \quad \quad \quad \quad \quad \quad \quad \quad \quad \quad \quad  \quad\;\; \\
= \frac{d D^{2}}{2\alpha_{T}} = \frac{d D^{2}}{2} \sqrt{T} \quad \quad \quad \quad \quad \quad \quad \quad \quad \quad \quad \quad \quad \quad \quad \quad \quad \quad \quad \quad \quad \quad \quad \quad \quad \quad \quad \quad \quad  \quad
\end{split}
\]

The inequality comes from the bounded diameter assumption, the non negativity of the norms, and the relationship between the infinity norm and the euclidean norm. The last equality is obtained using a property of the telescopic sequence. This completes the proof of the Lemma \ref{apsgd_convergence_lemma1}.
\end{lemma}

\begin{lemma}
\label{apsgd_convergence_lemma2}
From Equation \ref{equation_apsgd_convergence} the iterates of the error component, once evaluated in the optimal probabilities $\hat{p}_{t}$ will satisfy:
\[
\begin{split}
\mathbb{E}_{\hat{p}_{t}} \left[\,\hat{w}^{2}_{i_{t}} ||g_{i_{t}}||^{2}_{2}\,\Big|\,\theta_{t}\,\right]
= \mathbb{E}_{n}\left[||g_{i_{t}}||^{2}_{2}\right]-\text{Var}_{n}\left(||g_{i_{t}}||_{2}\right)
\end{split}
\]
\textit{Proof}
\[
\begin{split}
\mathbb{E}_{n}\left[||g_{i_{t}}||^{2}_{2}\right]-\text{Var}_{n}\left(||g_{i_{t}}||_{2}\right) 
\quad \quad \quad \quad \quad \quad \quad \quad \quad 
\quad \quad \quad \quad \quad \quad \quad \quad \quad \\
= \left(\mathbb{E}_{n}\left[||g_{i_{t}}||_{2}\right]\right)^{2} = \left(\sum^{n}_{i_{t}=1} \frac{||g_{i_{t}}||_{2}}{n}\right)^{2} 
\quad \quad \quad \quad \quad \quad \quad \;\; \\
= \left(\sum^{n}_{i_{t}=1} \frac{||g_{i_{t}}||_{2} p^{1/2}_{i_{t}} }{n p^{1/2}_{i_{t}}}\right)^{2} 
\leq \left(\sum^{n}_{i_{t}=1}\frac{||g_{i_{t}}||^{2}_{2}}{n^{2}p_{i_{}t}}\right) \left(\sum^{n}_{i_{t}=1}p_{i_{t}}\right) 
\quad \quad \quad \quad \quad \quad \quad \quad \\
= \left(\sum^{n}_{i_{t}=1}\frac{||g_{i_{t}}||^{2}_{2}}{n^{2}p_{i_{t}}}\right) 
= \mathbb{E}_{p_{t}} \left[\,{w}^{2}_{i_{t}} ||g_{i_{t}}||^{2}_{2}\,\Big|\,\theta_{t}\,\right] 
\quad \quad \quad \quad  \quad \quad \quad \quad
\end{split}
\]

The first equality comes from the variance definition, the first inequality comes from the non negativity of the norms, the second inequality is Cauchy-Schwarz. Finally we show that the lower bound is achievable by the optimal probabilities are $\hat{p}_{i_{t}} \propto ||\nabla f_{i_{t}}(\theta_{t})||_{2}$.
\[
\begin{split}
\mathbb{E}_{\hat{p}_{{t}}} \left[\,\hat{w}^{2}_{i_{t}} ||g_{i_{t}}||^{2}_{2}\,\Big|\,\theta_{t}\,\right] = \sum^{n}_{i_{t}} \frac{||g_{i_{t}}||^{2}_{2}}{n^{2}\hat{p}^{2}_{i_{t}}} \hat{p}_{i_{t}}  
\quad \quad \quad \quad \quad \quad \\
= \sum^{n}_{i_{t}} \frac{||g_{i_{t}}||^{2}_{2}}{n^{2}\left(\frac{||g_{i_{t}}||_{2}}{\sum^{n}_{i_{t}}||g_{i_{t}}||_{2}}\right)} = \left(\sum^{n}_{i_{t}=1} \frac{||g_{i_{t}}||_{2}}{n}\right)^{2} \end{split}
\]
\end{lemma}

\begin{lemma}
\label{apsgd_convergence_lemma3}
The total error will satisfy that:
\[
\begin{split}
\sum^{T}_{t=1} \frac{\alpha_{t}}{2} \mathbb{E}_{\hat{p}_{t}} \left[\,\hat{w}^{2}_{i_{t}} ||g_{i_{t}}||^{2}_{2}\,\Big|\,\theta_{t}\,\right]
\quad \quad \quad \quad \quad \quad \quad \quad \quad \quad \\
\leq  d G^{2} \left(\sqrt{T} - 1/2\right)
- \sum^{T}_{t=1} \text{Var}_{n}\left(||g_{i_{t}}||_{2}\right) 
\end{split}
\]
\textit{Proof}\\
Using the following bound of the hyper-harmonic sequence $\sum^{T}_{t=1}\frac{1}{\sqrt{t}} \leq (2\sqrt{T}-1)$:
\[
\begin{split}
\sum^{T}_{t=1} \frac{\alpha_{t}}{2} \mathbb{E}_{n} \left[\, ||g_{i_{t}}||^{2}_{2}\,\Big|\,\theta_{t}\,\right] \leq \sum^{T}_{t=1} \frac{\alpha_{t}}{2} ||G \odot \mathbbm{1} ||^{2}_{2} 
\quad \quad \quad \\
\leq \frac{d G^{2}}{2} \left(2 \sqrt{T} - 1\right)
\end{split}
\]
Therefore with the optimal probabilities $\hat{p}_{t}$:
\[
\begin{split}
\sum^{T}_{t=1} \frac{\alpha_{t}}{2} \mathbb{E}_{n} \left[\,\hat{w}^{2}_{i_{t}} ||g_{i_{t}}||^{2}_{2}\,\Big|\,\theta_{t}\,\right] \leq \sum^{T}_{t=1} \frac{\alpha_{t}}{2} ||G \odot \mathbbm{1} ||^{2}_{2} 
\quad \quad  \\
\leq  d G^{2} \left(\sqrt{T} - 1/2\right) - \sum^{T}_{t=1} \text{Var}_{n}\left(||g_{i_{t}}||_{2}\right) 
\;
\end{split}
\]

\end{lemma}
\vspace{.5cm}

Combining Lemma \ref{apsgd_convergence_lemma1}, Lemma \ref{apsgd_convergence_lemma2}, and \ref{apsgd_convergence_lemma3}  we finish the proof. \\

$\mathbf{Corollary \; \ref{apsgd_corollary}}$
\[
\begin{split}
R(\textsc{ap-SGD}) \leq d G^{2} (\sqrt{T}-1/2) 
\quad \quad \quad \quad \quad \quad \quad \quad\\
- \sum^{T}_{t=1} \text{Var}_{n}\left(||g_{i_{t}}||_{2}\right)+ \frac{d D^{2}}{2} \sqrt{T}
\end{split}
\]

\end{proof}

%% file: sections/section_appendix_th4.tex
The proof of Theorem \ref{dasgrad_convergence} assumes a convex differentiable objective function $f$, bounded diameter for the parameters, and bounded norm of the gradients. And will be given for any trajectory of probabilities $p_{t}$, Corollary \ref{dasgrad_corollary} addresses the optimal probabilities $\hat{p}_{t}$.

\begin{proof}
Analogous to the proof in Theorem \ref{apsgd_convergence} we build an upper bound of the expected regret using the convexity of the loss:
\[
f(\theta_{t}) - f(\theta^{*}) \leq \langle \,g_{t} ,\, \theta_{t}-\theta^{*} \rangle = \mathbb{E}_{n}\left[ \langle \,g_{i_{t}} ,\, \theta_{t}-\theta^{*} \rangle \right]
\]

While using \textsc{DASGrad} the update of the parameter will be given by the stochastic update dependent on the training example $i_{t}$ and the current parameter $\theta_{t}$:
\[
\begin{split}
\theta_{t+1} =  \Pi_{\Theta, \hat{V}^{1/2}_{i_{t}}} (\hat{\theta}_{t+1}) 
= \Pi_{\Theta, \hat{V}^{1/2}_{i_{t}}} (\theta_{t} - \alpha_{t} w_{i_{t}}\hat{V}^{-1/2}_{i_{t}} m_{i_{t}} ) \\
= \argminA_{\theta \in \Theta} ||\,\hat{V}^{1/4}_{i_{t}}(\theta_{t} - \alpha_{t} w_{i_{t}} \hat{V}^{-1/2}_{i_{t}} m_{i_{t}})\,||_{2}
\end{split}
\]

Then bound the expected regret of the algorithm, we use the fact that:
\[
\begin{split}
\hat{\theta}_{t+1}-\theta^{*}  =  (\theta_{t}-\theta^{*}) - \alpha_{t} w_{i_{t}}m_{i_{t}}/\sqrt{\hat{V}_{i_{t}}} 
\quad \quad \quad \quad \quad \quad \\
||\hat{V}^{1/4}_{i_{t}}(\hat{\theta}_{t+1}-\theta^{*})||^{2}_{2} = ||\hat{V}^{1/4}_{i_{t}}(\theta_{t}-\theta^{*})||^{2}_{2} \\
- 2\alpha_{t} w_{i_{t}} \langle \,m_{i_{t}} ,\, \theta_{t}-\theta^{*} \rangle + \alpha^{2}_{t} w^{2}_{i_{t}} ||\hat{V}^{-1/4}_{i_{t}} m_{i_{t}}||^{2}_{2} \\
 = ||\hat{V}^{1/4}_{i_{t}} (\theta_{t}-\theta^{*})||^{2}_{2} \\
- 2\alpha_{t} w_{i_{t}} \langle \,\beta_{1_{t}} m_{i_{t-1}}+(1-\beta_{1_{t}})g_{i_{t}} ,\, \theta_{t}-\theta^{*} \rangle \\
+ \alpha^{2}_{t} w^{2}_{i_{t}} ||\hat{V}^{-1/4}_{i_{t}} m_{i_{t}}||^{2}_{2}
\end{split}
\]

We identify the first three components as the potential, the immediate cost, now with extra terms associated to the moving average, and the error.\\

\newpage
\begin{lemma}
\label{extended_norm_projection}
For any $M \in S^{d}_{+}$ and convex feasible set $\Theta \subseteq \mathbb{R}^{d}$ with the projection operator $\Pi_{\Theta,M}$ let $u_{1} = \Pi_{\Theta,M}(z_{1})$ and $u_{2} = \Pi_{\Theta,M}(z_{2})$ then:
\[
||M^{1/2}\left(u_{1}-u_{2}\right)||_{2} \leq ||M^{1/2}\left(z_{1}-z_{2}\right)||_{2}
\]
\end{lemma}

Taking the expectation at time $t$, and using the extended norm reduction property of the projections from Lemma \ref{extended_norm_projection} we obtain the following inequality:
\[
\begin{split}
\mathbb{E}_{p_{t}}\left[||\hat{V}^{1/4}_{i_{t}}(\theta_{t+1}-\theta)||^{2}_{2}\Big|\theta_{t}\right] \leq 
\mathbb{E}_{p_{t}}\left[||\hat{V}^{1/4}_{i_{t}}(\theta_{t}-\theta^{*})||^{2}_{2}\Big|\theta_{t}\right]  
\\
- \mathbb{E}_{p_{t}}\left[\,2\alpha_{t} w_{i_{t}} \langle \,\beta_{1_{t}} m_{i_{t-1}}+(1-\beta_{1_{t}})g_{i_{t}} ,\, \theta_{t}-\theta^{*} \rangle \,\Big|\, \theta_{t}\,\right] \\
+ \alpha^{2}_{t} \mathbb{E}_{p_{t}} \left[\,w^{2}_{i_{t}} ||\hat{V}^{1/4}_{i_{t}}m_{i_{t}}||^{2}_{2}\,\Big|\,\theta_{t}\,\right]
\end{split}
\]

Since $w_{t}$ is the Radon-Nikodym derivative the interior product will be unbiased, then:
\[
\begin{split}
\mathbb{E}_{p_{t}}\left[\,||\hat{V}^{1/4}_{i_{t}}(\theta_{t+1}-\theta^{*})||^{2}_{2} \,\Big|\, \theta_{t}\,\right] \leq \quad \quad \quad \quad \quad \quad \quad \quad \quad \quad \quad \quad \quad \quad \quad \quad \quad \quad \\
\mathbb{E}_{p_{t}}\left[\,||\hat{V}^{1/4}_{i_{t}}(\theta_{t}-\theta^{*})||^{2}_{2}\,\Big|\, \theta_{t}\,\right] \quad \quad \quad \quad \quad \quad \quad \quad \\
- 2\alpha_{t} \mathbb{E}_{n}\left[ \langle \,\beta_{1_{t}} m_{i_{t-1}}+(1-\beta_{1_{t}})g_{i_{t}} ,\, \theta_{t}-\theta^{*} \rangle \right] \quad \quad \quad \quad \quad \quad \quad \quad \\
+ \alpha^{2}_{t} \, \mathbb{E}_{p_{t}} \left[\,w^{2}_{i_{t}} ||\hat{V}^{1/4}_{i_{t}}m_{i_{t}}||^{2}_{2}\,\Big|\,\theta_{t}\,\right] \quad \quad \quad \quad \quad \quad \quad \quad
\end{split}
\]

Finally rearranging the terms, summing until time $T$ and taking expectations of the complete Markovian process:
\begin{equation}\begin{split}
\label{equation_dasgrad_convergence}
R(\textsc{DASGrad}) \leq \quad \quad \quad \quad \quad \quad \quad \quad \quad \quad \quad \quad \quad \quad \quad \quad \quad \quad \quad \quad \quad \quad \quad \quad \quad \quad \quad
\quad \quad \quad \quad \\
\sum^{T}_{t=1} \frac{1}{2\alpha_{t}(1-\beta_{1_{t}})} \mathbb{E}_{p_{1:t}} \left[\,||\hat{V}^{1/4}_{i_{t}}(\theta_{t}-\theta^{*})||^{2}_{2} \right]
\quad \quad \quad \quad \quad \quad \quad \quad \quad \quad 
\quad \quad \quad \quad \quad \quad \quad \quad \quad \quad \\
- \sum^{T}_{t=1} \frac{1}{2\alpha_{t}(1-\beta_{1_{t}})}\mathbb{E}_{p_{1:t}} \left[\,||\hat{V}^{1/4}_{i_{t}}(\theta_{t+1}-\theta^{*})||^{2}_{2}\;\right] 
\quad \quad \quad \quad \quad \quad \quad \quad \quad 
\quad \quad \quad \quad \quad \quad  \\ 
+ \sum^{T}_{t=1}\frac{\alpha_{t}}{2(1-\beta_{1_{t}})} \mathbb{E}_{p_{1:t}} \left[\,w^{2}_{i_{t}} ||\hat{V}^{-1/4}_{i_{t}}m_{i_{t}}||^{2}_{2}\,\right] 
\quad \quad \quad \quad \quad \quad \quad \quad \quad \quad \quad \quad \quad \quad \quad  \\
+ \sum^{T}_{t=1} \frac{\beta_{1_{t}}}{2(1-\beta_{1_{t}})} \alpha_{t} ||\hat{V}^{-1/4}_{t}m_{t-1}||^{2}_{2} 
\quad \quad \quad \quad \quad \quad \quad \quad \quad \quad \quad \quad \quad \quad \quad \\
+ \sum^{T}_{t=1} \frac{\beta_{1_{t}}}{2\alpha_{t}(1-\beta_{1t})}
||\hat{V}^{1/4}_{t}(\theta_{t}-\theta^{*})||^{2}_{2} 
\quad \quad \quad \quad \quad \quad \quad \quad \quad \quad \quad \quad \quad \quad \quad 
\end{split}
\end{equation}

Last line is Cauchy-Schwarz and Young's inequality applied to the inner product of the extra terms associated with the moving average in the immediate cost.

\end{proof}

%% file: sections/section_appendix_c4_1.tex
\begin{proof}
The proof of Corollary \ref{dasgrad_convergence} is in the line of the improvements provided by Reddi et al. to the convergence proof of Kingma $\&$ Ba for \textsc{ADAM}, we adapt the arguments to the stochastic case. We assess separately each component of the expected regret from Equation \ref{equation_dasgrad_convergence}.

Lemma \ref{dasgrad_convergence_lemma1} addresses the potential, Lemma \ref{dasgrad_convergence_lemma2} the error, and Lemma \ref{dasgrad_convergence_lemma3} and Lemma \ref{dasgrad_convergence_lemma4} the moving average terms. The proof of Corollary \ref{dasgrad_corollary} is a consequence of all the previous Lemmas using the optimal probabilities while Corollary \ref{amsgrad_corollary} is the case with uniform probabilities.

Following the sequence $\{\theta_{t}\}^{T}_{t=1}$ of \textsc{DASGrad}, with step size $\alpha_{t} = \alpha/\sqrt{t}$, averaging parameters $\beta_{1} = \beta_{1_{1}}$ and $\beta_{1_{t}} \leq \beta_{1}$ for all $t\in[T]$ and $\gamma = \beta_{1}/\sqrt{\beta_2} < 1$. and bounded diameter $D$ for $\Theta$ and $||\nabla f_{i_{t}} (\theta)||_{\infty} \leq G$ for all $t \in [T]$ and $\theta \in \Theta$.

\begin{lemma}
\label{dasgrad_convergence_lemma1}
From Equation \ref{equation_dasgrad_convergence} the potential component will satisfy: 
\[
\begin{split}
\sum^{T}_{t=1} \frac{1}{2\alpha_{t}(1-\beta_{1_{t}})}  \mathbb{E}_{p_{1:t}} \left[||\hat{V}^{1/4}_{i_{t}}(\theta_{t}-\theta^{*})||^{2}_{2}-||\hat{V}^{1/4}_{i_{t}}(\theta_{t+1}-\theta^{*})||^{2}_{2}\right] 
\\
\leq \frac{D^{2}\sqrt{T}}{2\alpha(1-\beta_{1})} \mathbb{E}_{p_{1:T}}\left[||\hat{v}^{1/4}_{i_{T}}||^{2}_{2}\right]
\quad \quad \quad
\end{split}
\]

\textit{Proof}\\
As in Corollary \ref{apsgd_corollary} proof one can decompose the potential in the following manner:
\[
\begin{split}
\sum^{T}_{t=1} \frac{1}{2\alpha_{t}(1-\beta_{1_{t}})} \mathbb{E}_{p_{1:t}} \left[||\hat{V}^{1/4}_{i_{t}}(\theta_{t}-\theta^{*})||^{2}_{2}-||\hat{V}^{1/4}_{i_{t}}(\theta_{t+1}-\theta^{*})||^{2}_{2}\right] \quad \quad \quad \quad \quad \quad \quad \quad \quad \quad 
\quad \quad \quad \quad \quad \quad \quad \quad \quad \quad 
\quad \quad \quad \quad \quad \quad \quad \quad \quad \quad \\
\leq \frac{1}{2\alpha_{1}(1-\beta_{1})} \mathbb{E}_{p_{1}} \left[\,||\hat{V}^{1/4}_{i_{1}}(\theta_{1}-\theta^{*})||^{2}_{2} \,\right] \quad \quad \quad \quad \quad \quad \quad \quad \quad \quad \quad \quad \quad \quad \quad \quad \quad \quad \quad \quad \quad \quad \quad \quad \quad \quad \quad \quad \quad \quad \quad \quad \quad\\
- \frac{1}{2\alpha_{T}(1-\beta_{1})} \mathbb{E}_{p_{1:T}}\left[\,||\hat{V}^{1/4}_{i_{T}}(\theta_{T+1}-\theta^{*})||^{2}_{2}\,\right] \quad \quad \quad \quad \quad \quad \quad \quad \quad \quad \quad \quad \quad \quad \quad \quad \quad \quad \quad \quad \quad \quad \quad \quad \quad \quad \quad \quad \quad \quad \quad \quad \quad\\
+ \frac{1}{2(1-\beta_{1})} \sum^{T}_{t=2} \frac{1}{\alpha_{t}}\mathbb{E}_{p_{1:t}}\left[\,||\hat{V}^{1/4}_{i_{t}}(\theta_{t}-\theta^{*})||^{2}_{2}\,\right] \quad \quad \quad \quad \quad \quad \quad \quad \quad \quad \quad \quad \quad \quad \quad \quad \quad \quad \quad \quad \quad \quad \quad \quad \quad \quad \quad \quad \quad \quad \quad \quad \quad\\
- \frac{1}{2(1-\beta_{1})} \sum^{T}_{t=2} \frac{1}{\alpha_{t-1}}\mathbb{E}_{p_{1:t-1}}\left[\,||\hat{V}^{1/4}_{i_{t-1}}(\theta_{t-1}-\theta^{*})||^{2}_{2}\,\right] \quad \quad \quad \quad \quad \quad \quad \quad \quad \quad \quad \quad \quad \quad \quad \quad \quad \quad \quad \quad \quad \quad \quad \quad \quad \quad \quad \quad \quad \quad \quad \quad \quad\\
\leq \frac{1}{2\alpha_{1}(1-\beta_{1})} \mathbb{E}_{p_{1}} \left[\,||\hat{v}^{1/4}_{i_{1}} \odot D \odot \mathbbm{1}||^{2}_{2} \,\right] \quad \quad \quad \quad \quad \quad \quad \quad \quad \quad \quad \quad \quad \quad \quad \quad \quad \quad \quad \quad \quad \quad \quad \quad \quad \quad \quad \quad \quad \quad \quad \quad \quad \\
+ \frac{1}{2(1-\beta_{1})} \sum^{T}_{t=2} \frac{1}{\alpha_{t}} \mathbb{E}_{p_{1:t}}\left[\,||\hat{v}^{1/4}_{i_{t}} \odot D \odot \mathbbm{1}||^{2}_{2}\,\right] \quad \quad \quad \quad \quad \quad \quad \quad \quad \quad \quad \quad \quad \quad \quad \quad \quad \quad \quad \quad \quad \quad \quad \quad \quad \quad \quad \quad \quad \quad \quad \quad \quad\\
- \frac{1}{2(1-\beta_{1})} \sum^{T}_{t=2} \frac{1}{\alpha_{t-1}} \mathbb{E}_{p_{1:t-1}}\left[\,||\hat{v}^{1/4}_{i_{t-1}} \odot D \odot \mathbbm{1}||^{2}_{2}\,\right] \quad \quad \quad \quad \quad \quad \quad \quad \quad \quad \quad \quad \quad \quad \quad \quad \quad \quad \quad \quad \quad \quad \quad \quad \quad \quad \quad \quad \quad \quad \quad \quad \quad
\end{split}
\]

\newpage
Finally,
\[
\begin{split}
\sum^{T}_{t=1} \frac{1}{2\alpha_{t}(1-\beta_{1_{t}})} \mathbb{E}_{p_{1:t}} \left[||\hat{V}^{1/4}_{i_{t}}(\theta_{t}-\theta^{*})||^{2}_{2}\right]
\quad \quad \quad \quad \quad \quad \quad \quad
\quad \quad \quad \quad \quad \quad \quad \quad 
\quad \quad \quad \quad \quad \quad \quad \quad 
\quad \quad \quad \quad \quad \quad \quad \quad
\quad \quad \quad \quad \quad \quad \quad \quad
\quad \quad \quad \quad \quad \quad \quad \quad
\quad \quad \quad \quad \quad \\
-\sum^{T}_{t=1} \frac{1}{2\alpha_{t}(1-\beta_{1_{t}})}
\mathbb{E}_{p_{1:t}}\left[||\hat{V}^{1/4}_{i_{t}}(\theta_{t+1}-\theta^{*})||^{2}_{2}\right]
\quad \quad \quad \quad \quad \quad \quad \quad
\quad \quad \quad \quad \quad \quad \quad \quad 
\quad \quad \quad \quad \quad \quad \quad \quad 
\quad \quad \quad \quad \quad \quad \quad \quad
\quad \quad \quad \quad \quad \quad \quad \quad 
\quad \quad \quad \quad \quad \quad \quad \quad \\
\leq 
\frac{D^2}{2(1-\beta_1)} 
\left(
\frac{1}{\alpha_{1}}\mathbb{E}_{p_{1}}\left[||\hat{v}^{1/4}_{i_{1}}||^{2}_{2}\right] + \right. 
\quad \quad \quad \quad \quad \quad \quad \quad
\quad \quad \quad \quad \quad \quad \quad \quad 
\quad \quad \quad \quad \quad \quad \quad \quad 
\quad \quad \quad \quad \quad \quad \quad \quad
\quad \quad \quad \quad \quad \quad \quad \quad
\quad \quad \quad \quad \quad \quad \quad \quad
\quad \quad \quad \quad \quad \quad \quad \quad
\\ 
\left. \sum^{T}_{t=2} \left(\frac{1}{\alpha_{t}} - \frac{1}{\alpha_{t-1}} \right)\mathbb{E}_{p_{t}}\left[||\hat{v}^{1/4}_{i_{t}}||^{2}_{2}\right] 
\right)
\quad \quad \quad \quad \quad \quad \quad \quad
\quad \quad \quad \quad \quad \quad \quad \quad 
\quad \quad \quad \quad \quad \quad \quad \quad 
\quad \quad \quad \quad \quad \quad \quad \quad
\quad \quad \quad \quad \quad \quad \quad \quad 
\quad \quad \quad \quad \quad \quad \quad \quad\\
= \frac{D^2}{2\alpha_{T}(1-\beta_1)} \mathbb{E}_{p_{1:T}}\left[\,||\hat{v}^{1/4}_{i_{T}}||^{2}_{2}\,\right] \quad \quad \quad \quad \quad \quad \quad 
\quad \quad \quad \quad \quad \quad \quad \quad 
\quad \quad \quad \quad \quad \quad \quad \quad 
\quad \quad \quad \quad \quad \quad \quad \quad 
\quad \quad \quad \quad \quad \quad \quad \quad 
\quad \quad \quad \quad \quad \quad \quad \quad 
\quad
\end{split}
\]
The first inequality comes from rearranging and the definition of $\beta_{1_{t}}$, the second inequality comes from the bounded diameter assumption applied to each entry of $\theta_{t}-\theta^{*}$ and using the Hadamard's product to represent the original matrix multiplication, the third inequality\footnote{The third inequality is of particular importance since Reddi et al. showed that it is one of the main flaws in the convergence analysis of \textsc{ADAM} and \textsc{RMSProp}, and provided a simple fix to the adaptive moment methods that guarantees the non increasing property needed to achieve the telescopic sequence upper bound.} comes from the definition of $\hat{v}_{i_{t}} = \text{max}(\hat{v}_{i_{t-1}}, v_{i_{t}})$, the last equality comes from the property of the telescopic sequence. This completes the proof of Lemma \ref{dasgrad_convergence_lemma1}.
\end{lemma}

\begin{lemma}
\label{dasgrad_convergence_lemma2}
From Equation \ref{equation_dasgrad_convergence} the error component, once evaluated in the optimal probabilities $\hat{p}_{t}$ will satisfy:
\[
\begin{split}
\mathbb{E}_{\hat{p}_{t}} \left[\,\hat{w}^{2}_{i_{t}} ||\hat{V}^{1/4}_{i_{t}}m_{i_{t}}||^{2}_{2}\,\Big|\,\theta_{t}\,\right] 
\quad \quad \quad \quad \quad \quad \quad \quad \quad \quad \quad \quad \\
= \mathbb{E}_{n} \left[\, ||\hat{V}^{1/4}_{i_{t}}m_{i_{t}}||^{2}_{2}\,\Big|\,\theta_{t}\,\right] 
- \text{Var}_{n}\left(||\hat{V}^{1/4}_{i_{t}}m_{i_{t}}||_{2}\right)
\end{split}
\]
\textit{Proof}\\
The proof follows analogous arguments to Lemma \ref{apsgd_convergence_lemma2}, creating a lower bound with Cauchy-Schwarz and showing that it is achievable with the optimal probabilities $\hat{p}_{i_{t}} \propto ||\hat{V}^{1/4}_{i_{t}}m_{i_{t}}||_{2}$.
\end{lemma}

\begin{lemma}
\label{dasgrad_convergence_lemma4}
The first component of the extra terms associated with the moving average in Equation \ref{equation_dasgrad_convergence} will satisfy:
\[
\begin{split}
\sum^{T}_{t=1} \frac{\alpha_{t}\beta_{1_{t}}}{2\alpha_{t}(1-\beta_{1_{t}})} ||\hat{V}^{-1/4}_{t} m_{t-1} ||^{2}_{2} \leq \quad \quad \quad \quad \quad \quad \quad \quad \\
\frac{\alpha G d}{2\alpha(1-\beta_{1})^{3} \sqrt{(1-\beta_{2})} (1-\gamma)}
\end{split}
\]
\newpage
\textit{Proof} \\
Following very similar arguments as those from Lemma \ref{dasgrad_convergence_lemma3}, we can get:
\[
\begin{split}
\sum^{T}_{t=1} \frac{\alpha_{t}\beta_{1_{t}}}{2\alpha_{t}(1-\beta_{1_{t}})} ||\hat{V}^{-1/4}_{t} m_{t-1} ||^{2}_{2} \leq \quad \quad \quad \quad \quad \quad \quad \quad \quad \quad \quad \quad \quad \quad \quad \quad \quad \quad \quad \quad \\
 \frac{\alpha}{2(1-\beta_{1})^{2} \sqrt{(1-\beta_{2})}(1-\gamma)} \sum^{T}_{t=1} \beta^{T-t}_{1} ||g_{t}||_{1} \quad \quad \quad \quad \quad \quad \quad \quad \quad \quad \quad \quad\\
\leq \frac{\alpha}{2(1-\beta_{1})^{2} \sqrt{(1-\beta_{2})}(1-\gamma)} \sum^{T}_{t=1} \beta^{T-t}_{1} ||G \odot \mathbbm{1} ||_{1} \quad \quad \quad \quad \quad \quad \quad \quad \quad \quad \quad \quad \\
\leq \frac{\alpha G d}{2\alpha(1-\beta_{1})^{3} \sqrt{(1-\beta_{2})} (1-\gamma)} \quad \quad \quad \quad \quad \quad \quad \quad \quad \quad \quad \quad
\end{split}
\]

This completes the proof of Lemma \ref{dasgrad_convergence_lemma4}.
\end{lemma}

\begin{lemma}
\label{dasgrad_convergence_lemma5}
To finish the second component of the extra terms associated with the moving average in Equation \ref{equation_dasgrad_convergence}, will satisfy:
\[
\begin{split}
    \sum^{T}_{t=1} \frac{\beta_{1_{t}}}{2\alpha_{t}(1-\beta_{1_{t}})} ||\hat{V}^{1/4}_{t} (\theta_{t}-\theta^{*})||^{2}_{2} \leq \quad \quad \quad \quad \quad \quad \quad \\ \frac{D^{2}}{2\alpha(1-\beta_{1})} \sum^{T}_{t=1} \sqrt{t} \beta^{T-t}_{1} ||\hat{v}^{1/4}_{t}||^{2}_{2} 
\end{split}
\]

\textit{Proof}
\[
\begin{split}
\sum^{T}_{t=1} \frac{\beta_{1_{t}}}{2\alpha_{t}(1-\beta_{1_{t}})} ||\hat{V}^{1/4}_{t} (\theta_{t}-\theta^{*})||^{2}_{2} \quad \quad \quad \quad \quad \quad \quad \quad \quad\\\leq \sum^{T}_{t=1} \frac{\beta_{1_{t}}}{2\alpha_{t}(1-\beta_{1_{t}})} ||\hat{v}^{1/4}_{t} \odot D \odot \mathbbm{1}||^{2}_{2} \\
= \frac{D^2}{2 \alpha} \sum^{T}_{t=1} \sqrt{t} \frac{\beta_{1_{t}}}{(1-\beta_{1_{t}})} ||\hat{v}^{1/4}_{t} ||^{2}_{2} \quad \quad \\
\leq \frac{D^2}{2 \alpha (1-\beta_{1})} \sum^{T}_{t=1} \sqrt{t} \beta^{T-t}_{1}  ||\hat{v}^{1/4}_{t} ||^{2}_{2}\quad 
\end{split}
\]
This completes the proof of Lemma \ref{dasgrad_convergence_lemma5}.
\end{lemma}

\newpage
\begin{lemma}
\label{dasgrad_convergence_lemma3}
From Equation \ref{equation_dasgrad_convergence} the error component, once evaluated in the optimal probabilities $\hat{p}_{t}$, and the total error will satisfy that:
\[
\begin{split}
\sum^{T}_{t=1} \frac{\alpha_{t}}{2(1-\beta_{1_{t}})} \mathbb{E}_{\hat{p}_{t}} \left[\,\hat{w}^{2}_{i_{t}} ||\hat{V}^{1/4}_{i_{t}}m_{i_{t}}||^{2}_{2}\,\Big|\,\theta_{t}\,\right]
\leq 
\quad \quad \quad \\
\frac{\alpha \sqrt{1+\log(T)}}{2(1-\beta_{1})^2\sqrt{(1-\beta_2)}(1-\gamma)} \sum^{d}_{h=1} ||\;\bar{|\,g\,|}_{1:T,h}\;||_{2} \\ -\sum^{T}_{t=1} \text{Var}_{n}\left(||\hat{V}^{1/4}_{i_{t}}m_{i_{t}}||_{2}\right)
\end{split}
\]

\textit{Proof} \\
For Lemma \ref{dasgrad_convergence_lemma3} we follow Kingma $\&$ Ba, for every element at time $t$ of the error component:
\begin{flalign*}
\aalpha \mathbb{E}_{n}\left[||\hat{V}^{-1/4}_{i_{t}} m_{i_{t}}||^{2}_{2} \right]
\leq \\
\aalpha \mathbb{E}_{n}\left[||V^{-1/4}_{i_{t}} m_{i_{t}}||^{2}_{2} \right] \\
=
\aalpha \mathbb{E}_{n} \left[\frac{||\Sigma^{t}_{\tau=1} \text{\scriptsize $\beta_{1}(t)_{\tau}$} g_{i_{\tau}}||^{2}_{2} }{\sqrt{v_{i_{t}}}}\right] \\
= 
\aalpha \mathbb{E}_{n} \left[ \frac{||\Sigma^{t}_{\tau=1} \text{\scriptsize $\beta_{1}(t)^{1/2}_{\tau} \beta_{1}(t)^{1/2}_{\tau}$} g_{i_{\tau}}||^{2}_{2}}{\sqrt{v_{i_{\tau}}}} \right] \\
\leq
\aalpha \mathbb{E}_{n} \left[ \left(\sum^{t}_{\tau=1} \beta_{1}(t)_{\tau} \right)\left(\sum^{t}_{\tau=1} \beta_{1}(t)_{\tau} \frac{ ||g_{i_{\tau}}||^{2}_{2} }{\sqrt{v_{i_{\tau}}}} \right)\right] \\
\leq
\frac{\alpha_{t}}{2(1-\beta_{1})} \mathbb{E}_{n} \left[ \left(\sum^{t}_{\tau=1} \beta^{t-\tau}_{1}\right)\left(\sum^{t}_{\tau=1} \beta^{t-\tau}_{1}\frac{ ||g_{i_{\tau}}||^{2}_{2} }{\sqrt{v_{i_{\tau}}}} \right)\right] \\
\leq 
\frac{\alpha_{t}}{2(1-\beta_{1})^2} \mathbb{E}_{n} \left[\,\left(\sum^{t}_{\tau=1} \beta^{t-\tau}_{1}\frac{||g_{i_{\tau}}||^{2}_{2}}{ \sqrt{(1-\text{\scriptsize $\beta_{2}$})\Sigma^{t}_{\tau=1} \text{\scriptsize $\beta^{t-\tau}_{2}$} g^{2}_{i_{\tau}}}} \right)\right] \\
\leq 
\frac{\alpha_{t}}{2(1-\beta_{1})^2\sqrt{(1-\beta_2)}} \mathbb{E}_{n}\left[\left(\sum^{t}_{\tau=1} \frac{\beta^{t-\tau}_{1}}{\sqrt{\beta^{t-\tau}_{2}}}\frac{||g_{i_{\tau}}||^{2}_{2}}{|g_{i_{\tau}}|} \right)\right] \\
=
\frac{\alpha }{2(1-\beta_{1})^2\sqrt{(1-\beta_2)}} \mathbb{E}_{n}\left[ \frac{1}{\sqrt{t}}\left(\sum^{t}_{\tau=1} \gamma^{t-\tau} ||g_{i_{\tau}}||_{1} \right)\right]
\end{flalign*}

The first inequality follows the definition of the auxiliary vectors $\hat{v}_{t}=\text{max}(\hat{v}_{t-1}, v_{t})$, the second inequality comes from the non negativity of $\beta_{1}(t)_{\tau}$. The third and fourth inequality comes from the decreasing property $\beta_{1} \leq \beta_{1_{1}}$ and $\beta_{1_{t}} \leq \beta_{1_{1}}$ and the property of the geometric sequence. The fifth inequality comes from the non negativity of $\beta_{2}$ and $||g_{i_{\tau}}||^{2}_{2}$, the last equality uses the definition of the step size. \\
Finally using induction one can show that:
\begin{equation}
\label{equation_dasgrad_lemma3}
\begin{split}
\sum^{T}_{t=1} \frac{\alpha_{t}}{2(1-\beta_{1_{t}})} \mathbb{E}_{n}\left[\,||\hat{V}^{1/4}_{i_{t}}m_{i_{t}}||^{2}_{2} \,\right] 
\leq \quad \quad \quad \quad \quad \quad 
\quad \quad \quad \quad \quad \quad \quad 
\quad \quad \quad \quad \quad \quad \quad 
\quad \quad \\
\sum^{T}_{t=1} \frac{\alpha}{2(1-\beta_{1})^2\sqrt{(1-\beta_2)}} \mathbb{E}_{n}\left[\,\left(\sum^{T}_{\tau=t} \frac{\gamma^{t-\tau}}{\sqrt{\tau}} ||g_{i_{\tau}}||_{1} \right)\right] 
\quad \quad \quad \quad \quad \quad \quad 
\quad \quad \quad \quad \quad \quad
\end{split}
\end{equation}
Continuing the proof of Lemma \ref{dasgrad_convergence_lemma3}, let $k=\frac{\alpha}{2(1-\beta_{1})^2\sqrt{(1-\beta_2)}}$, from Equation \ref{equation_dasgrad_lemma3} we have that:
\begin{flalign*}
\begin{split}
\sum^{T}_{t=1} \frac{\alpha_{t}}{2(1-\beta_{1_{t}})} \mathbb{E}_{n}\left[\,||V^{1/4}_{i_{t}}m_{i_{t}}||^{2}_{2} \,\right] 
\leq 
\quad \quad \quad \quad \quad \quad \quad 
\quad \quad \quad \quad \quad \quad \quad \quad \quad \quad \quad \quad \quad \quad\quad 
\quad \quad \quad \quad \\
\sum^{T}_{t=1} k \; \mathbb{E}_{n}\left[ \frac{1}{\sqrt{t}} \left(\sum^{T}_{\tau=t} \gamma^{t-\tau} ||g_{i_{\tau}}||_{1} \right)\right] \quad \quad \quad \quad \quad \quad \quad \quad \quad \quad \quad \quad\quad \quad \quad \quad \quad\\
= 
k \; \left( \sum^{T}_{t=1} \sum^{n}_{i_{t}=1}\frac{1}{n} ||g_{i_{t}}||_{1} \left(\sum^{T}_{\tau=t} \frac{\gamma^{t-\tau}}{\sqrt{\tau}} \right) \right) \quad \quad \quad \quad \quad \quad\quad \quad \quad \quad \quad \quad \quad \quad \quad \quad \quad\\
\leq 
k \left( \sum^{T}_{t=1} \sum^{n}_{i_{t}=1}\frac{1}{n} ||g_{i_{t}}||_{1} \left(\sum^{T}_{\tau=t} \frac{\gamma^{t-\tau}}{\sqrt{t}} \right) \right) \quad \quad \quad \quad \quad \quad \quad \quad \quad \quad \quad \quad \quad \quad \quad\quad\quad\\
=
k \left(\sum^{T}_{t=1} \sum^{n}_{i_{t}=1}\frac{1}{n} ||g_{i_{t}}||_{1} \frac{1}{\sqrt{t}} \left(\sum^{T}_{\tau=t} \gamma^{t-\tau} \right)\right) \quad \quad\quad \quad \quad \quad \quad \quad \quad \quad \quad \quad \quad \quad \quad \quad\quad\\
\leq 
\frac{k}{(1-\gamma)} \left(\sum^{T}_{t=1} \left(\sum^{n}_{i_{t}=1}\frac{1}{n} ||g_{i_{t}}||_{1}\right) \left(\frac{1}{\sqrt{t}}\right) \right) \quad \quad \quad \quad \quad \quad \quad \quad\quad \quad \quad \quad \quad \quad \quad \quad \quad\\
=  
\frac{k}{(1-\gamma)} \sum^{d}_{h=1} \left(\sum^{T}_{t=1} \left(\sum^{n}_{i_{t}=1}\frac{1}{n} |g_{i_{t},h}| \right) \left(\frac{1}{\sqrt{t}}\right) \right)\quad \quad \quad \quad \quad \quad \quad \quad \quad \quad \quad \quad\quad \quad \quad \quad \quad \\
\leq 
\frac{k}{(1-\gamma)}  \sum^{d}_{h=1} \left( \sqrt{\sum^{T}_{t=1} \left(\sum^{n}_{i_{t}=1}\frac{1}{n} |g_{i_{t},h}|\right)^2} \sqrt{\sum^{T}_{t} \frac{1}{t}} \right)  \quad \quad \quad \quad \quad \quad \quad \quad  \quad \quad \quad \quad \quad \quad \quad \quad \quad \\
\leq 
\frac{k \sqrt{1+\log(T)}}{(1-\gamma)} \sum^{d}_{h=1} \sqrt{\sum^{T}_{t=1} \left(\sum^{n}_{i_{t}=1}\frac{1}{n} |g_{i_{t},h}|\right)^2}\quad \quad \quad \quad \quad \quad  \quad \quad \quad \quad \quad \quad \quad \quad \quad \quad \quad
\end{split}
\end{flalign*}
The first equality comes from a change of indexes, the second inequality is an upper bound for the arithmetic sequence that begins at $t$, the third inequality is an upper bound for the geometric sequence, the fourth inequality comes is an application of Cauchy-Schwarz inequality, finally the fifth inequality is an upper bound for the arithmetic sequence. 
\[
\begin{split}
\sum^{T}_{t=1} \mathbb{E}_{\hat{p}_{t}} \left[\,\hat{w}^{2}_{i_{t}} ||\hat{V}^{1/4}_{i_{t}}m_{i_{t}}||^{2}_{2}\,\Big|\,\theta_{t}\,\right] = \quad \quad \quad \quad \quad \quad \quad \quad \quad \quad \quad \quad \quad \quad \quad \quad \\
\sum^{T}_{t=1} \mathbb{E}_{n} \left[ ||\hat{V}^{1/4}_{i_{t}}m_{i_{t}}||^{2}_{2}\,\Big|\,\theta_{t}\,\right] 
- \sum^{T}_{t=1} \text{Var}_{n}\left(||\hat{V}^{1/4}_{i_{t}}m_{i_{t}}||_{2}\right) \quad \quad \quad \quad \quad \quad \quad \quad \\
\leq \frac{\alpha \sqrt{1+\log(T)}}{2(1-\beta_{1})^2\sqrt{(1-\beta_2)}(1-\gamma)} \sum^{d}_{h=1} ||\bar{|\,g\,|}_{1:T,h}\;||_{2} \quad \quad \quad \quad \quad \quad \quad \quad \\
- \sum^{T}_{t=1} \text{Var}_{n}\left(||\hat{V}^{1/4}_{i_{t}}m_{i_{t}}||_{2}\right) \quad \quad \quad \quad \quad \quad \quad \quad
\end{split}
\]

With the optimal probabilities $\hat{p}_{i_{t}}$, we complete the proof of Lemma \ref{dasgrad_convergence_lemma3}.
\end{lemma}

\vspace{1cm}

Finally we can combine the results from Lemma \ref{dasgrad_convergence_lemma1} to \ref{dasgrad_convergence_lemma5} and obtain the following bound for the expected regret of the general double adaptive algorithms:
\vspace{1cm}

$\mathbf{Corollary \; \ref{dasgrad_corollary}}$

\[\begin{split}
R(\textsc{DASGrad})_{T} 
\leq \frac{\alpha \sqrt{1+\log(T)}}{2(1-\beta_{1})^2\sqrt{(1-\beta_2)}(1-\gamma)} \sum^{d}_{h=1} ||\;\bar{|\,g\,|}_{1:T,h}\;||_{2} \quad \quad \quad \quad \quad \quad \quad \quad \quad \quad \quad \quad \quad \quad \quad \quad \quad \quad\\
-\sum^{T}_{t=1} \text{Var}_{n}\left(||\hat{V}^{1/4}_{i_{t}}m_{i_{t}}||_{2}\right)
+  \frac{D^{2}\sqrt{T}}{2\alpha(1-\beta_{1})} \mathbb{E}_{\hat{p}_{1:T}}\left[||\hat{v}^{1/4}_{i_{T}}||^{2}_{2}\right] \quad \quad \quad \quad \quad \quad \quad \quad \quad \quad \quad \quad \quad \quad \quad \quad \quad \quad\\\frac{\alpha G d}{2\alpha(1-\beta_{1})^{3} \sqrt{(1-\beta_{2})} (1-\gamma) } \quad \quad \quad \quad \quad \quad \quad \quad \quad \quad \quad \quad \quad \quad \quad \quad \quad \quad
\\ + \frac{D^{2}}{2\alpha(1-\beta_{1})} \sum^{T}_{t=1} \sqrt{t} \beta^{T-t}_{1} ||\hat{v}^{1/4}_{t}||^{2}_{2}\quad \quad \quad \quad \quad \quad \quad \quad \quad \quad \quad \quad \quad \quad \quad \quad \quad \quad
\end{split}
\]

\end{proof}